\documentclass{article} 
\usepackage{iclr2027_conference,times}

\usepackage{booktabs} 
\usepackage[utf8]{inputenc} 
\usepackage[T1]{fontenc}    
\usepackage{microtype}      
\usepackage{hyperref}
\usepackage{url}
\usepackage{amsmath}
\usepackage{algorithm}
\usepackage{algpseudocode}
\usepackage{amsthm}
\usepackage{bm}
\usepackage{graphicx}
\usepackage{pifont}
\usepackage{makecell}
\usepackage{amssymb}
\usepackage{mathtools}
\usepackage{nicefrac}
\usepackage{amsfonts} 
\usepackage{calc}
\usepackage{subcaption}
\usepackage{xcolor} 
\usepackage{xspace} 
\usepackage[disable]{todonotes}

\usepackage{enumitem}

\renewcommand{\eqref}[1]{Equation~(\ref{#1})}
\newcommand{\tabref}[1]{Table~\ref{#1}}
\newcommand{\appref}[1]{Appendix~\ref{#1}}

\hypersetup{
colorlinks=true,
linkcolor={red!50!black},
filecolor=magenta,      
urlcolor={blue!80!black},
citecolor={blue!50!black},
pdftitle={Multi-objective Hyperparameter Optimization
with Expert Priors}
}

\newcommand\ie{\textit{i.e.}}
\newcommand{\eg}{\textit{e.g.}}

\newcommand{\sota}{state-of-the-art}
\newcommand{\topk}{\textit{top\_k}}

\newcommand{\cmark}{\ding{51}}%
\newcommand{\xmark}{\ding{53}}%

\newcommand{\rqa}{\textbf{RQ1}}
\newcommand{\rqb}{\textbf{RQ2}}
\newcommand{\rqc}{\textbf{RQ3}}
\newcommand{\rqd}{\textbf{RQ4}} 
\newcommand{\rqe}{\textbf{RQ5}} 
\newcommand{\rqf}{\textbf{RQ6}}

\newcommand{\init}{initial design}

\newcommand{\ova}{overall}

\newcommand{\allg}{all priors good}

\newcommand{\allb}{all priors bad}

\newcommand{\mxd}{mixed priors}

\newcommand{\eps}{$\epsilon$}

\newcommand{\minfid}{z_{min}}
\newcommand{\maxfid}{z_{max}}

\newcommand{\nfevals}{20}

\newcommand{\algo}{PriMO}

\newcommand{\hp}{\text{hyperparameter}}

\newcommand{\dl}{\text{DL}}
\newcommand{\dllong}{Deep Learning}
\newcommand{\hpo}{\text{HPO}}

\newcommand{\so}{\text{SO}}
\newcommand{\solong}{single-objective}

\newcommand{\soolong}{single-objective optimization}
\newcommand{\mo}{\text{MO}}
\newcommand{\molong}{multi-objective}

\newcommand{\moolong}{multi-objective optimization}
\newcommand{\mf}{\text{MF}}
\newcommand{\mflong}{multi-fidelity}

\newcommand{\mfolong}{multi-fidelity optimization}
\newcommand{\momf}{\text{MOMF}}
\newcommand{\momflong}{multi-objective multi-fidelity}

\newcommand{\mohpolong}{multi-objective hyperparameter optimization}

\newcommand{\mfpbench}{\texttt{mf-prior-bench}\ }
\newcommand{\yahpogym}{Yahpo-Gym\ }

\newcommand{\hyperbo}{HyperBO\ }
\newcommand{\pdone}{PD1}
\newcommand{\lcbench}{LCBench}

\newcommand{\cifar}{cifar-100}
\newcommand{\imagenet}{imagenet}
\newcommand{\translate}{translate-wmt-xformer}
\newcommand{\lmbt}{lm1b-transformer}

\newcommand{\lcba}{LCBench-126026}
\newcommand{\lcbb}{LCBench-146212}
\newcommand{\lcbc}{LCBench-168330}
\newcommand{\lcbd}{LCBench-168868}

\newcommand{\RS}{\text{RS}}
\newcommand{\RSlong}{Random Search}

\newcommand{\hb}{\text{HB}} 
\newcommand{\hblong}{HyperBand}

\newcommand{\sh}{\text{SH}}
\newcommand{\shlong}{Successive Halving}

\newcommand{\bo}{\text{BO}}
\newcommand{\bolong}{Bayesian Optimization}

\newcommand{\bohb}{\text{BOHB}}

\newcommand{\asha}{\text{ASHA}}

\newcommand{\pb}{Priorband}

\newcommand{\pibo}{$\pi$BO}

\newcommand{\RSprior}{RS + Prior}
\newcommand{\RSpriorlong}{Random Search + Prior}

\newcommand{\mobo}{\text{MO-BO}}
\newcommand{\mobolong}{Multi-objective Bayesian Optimization}

\newcommand{\borw}{\text{BO+RW}}
\newcommand{\borwlong}{Bayesian Optimization with Random Weights}

\newcommand{\nsgaii}{\text{NSGA-II}}
\newcommand{\nsgaiilong}{Non-dominated Sorting Genetic Algorithm}

\newcommand{\momfbo}{\text{MOMF-BO}}
\newcommand{\momfbolong}{Model-based Multi-objective Multi-fidelity}

\newcommand{\smsemoa}{\text{SMS-EMOA}}
\newcommand{\smsemoalong}{S-Metric Selection Evolutionary Multi-objective Optimization Algorithm}

\newcommand{\parego}{\text{ParEGO}}

\newcommand{\hbrw}{\text{HB+RW}} 
\newcommand{\hbrwlong}{HyperBand with Random Weights}

\newcommand{\moasha}{\text{MOASHA}}
\newcommand{\moashalong}{Multi-objective Aynchronous Successive Halving}

\newcommand{\priormoasha}{MOASHA + Prior}

\newcommand{\mohblong}{MO-HyperBand}

\newcommand{\piborw}{$\pi$BO+RW}
\newcommand{\piborwlong}{$\pi$BO with random weights}

\newcommand{\lnehvi}{LogNEHVI}
\newcommand{\lnehvilong}{Log Noisy Expected Hypervolume Improvement}

\newcommand{\motpe}{MO-TPE}

\newcommand{\mopb}{MO-Priorband}

\newcommand{\acqfunc}{acquisition function}

\newcommand{\ndslong}{non-dominated sorting}

\newcommand{\cdslong}{crowding-distance sort}

\newcommand{\epsnet}{$\epsilon$-net}
\newcommand{\epsnetlong}{\texttt{epsilon-net}}

\newcommand{\gp}{GP}

\newcommand{\ei}{\text{EI}}
\newcommand{\eilong}{Expected Improvement}

\newcommand{\hv}{\text{HV}}
\newcommand{\hvlong}{Hypervolume}

\newcommand{\hvi}{\text{HVI}}
\newcommand{\hvilong}{Hypervolume Improvement}

\newcommand{\ehvi}{\text{EHVI}}
\newcommand{\ehvilong}{Expected Hypervolume Improvement}

\newcommand{\neps}{NePS}

\newcommand{\botorch}{BoTorch}

\newcommand{\nvg}{Nevergrad}

\newcommand{\smac}{SMAC3}

\newcommand{\hpogluett}{\texttt{hpoglue}}

\newcommand{\synetune}{Syne Tune}

\newcommand{\answerYes}[1]{{\textcolor{answerYesColor}{[Yes]}\xspace#1}}

\newcommand{\answerNA}[1]{{\textcolor{answerNAColor}{[N/A]}\xspace#1}}

\definecolor{answerYesColor}{RGB}{31,120,180}
\definecolor{answerNoColor}{RGB}{227,26,28}
\definecolor{answerNAColor}{RGB}{32, 32, 32}
\definecolor{answerTODOColor}{RGB}{255,127,0}

\usepackage{amsmath,amsfonts,bm}

\def\figref#1{figure~\ref{#1}}
\def\Figref#1{Figure~\ref{#1}}

\def\Secref#1{Section~\ref{#1}}

\def\eqref#1{equation~\ref{#1}}

\def\1{\bm{1}}

\DeclareMathAlphabet{\mathsfit}{\encodingdefault}{\sfdefault}{m}{sl}
\SetMathAlphabet{\mathsfit}{bold}{\encodingdefault}{\sfdefault}{bx}{n}

\title{Multi-objective Hyperparameter Optimization
with Expert Priors}

\author{%
\textbf{Soham Basu}\textsuperscript{1,2}\thanks{Most of the work was conducted while the author was at the University of Freiburg.}\quad
\textbf{Nastaran Alipour}\textsuperscript{2,3}\quad
\textbf{Frank Hutter}\textsuperscript{2,4,5}\quad
\textbf{Danny Stoll}\textsuperscript{2}\\
\textsuperscript{1}University of Tübingen\quad
\textsuperscript{2}University of Freiburg\quad
\textsuperscript{3}Zuse School ELIZA\quad \\
\textsuperscript{4}Prior Labs\quad
\textsuperscript{5}ELLIS Institute Tübingen\\
{\small
\texttt{soham.basu@uni-tuebingen.de}\quad
\texttt{\{alipourn,fh,stolld\}@cs.uni-freiburg.de}}%
}

\iclrfinalcopy 
\begin{document}

\fancyhead{}
\renewcommand{\headrulewidth}{0pt}

\maketitle

\begin{abstract}
While \dllong{} (\dl{}) experts often have prior knowledge about which \hp{} settings yield strong performance, only few Hyperparameter Optimization (\hpo{}) algorithms can leverage such prior knowledge and none incorporate priors over multiple objectives. As \dl{} practitioners often need to optimize not just one but many objectives,
this is a blind spot in the algorithmic landscape of \hpo{}.
To address this shortcoming, we propose the novel problem setting of multi-objective \hpo{} with expert priors and introduce \algo{}, the first \hpo{} algorithm that can integrate user beliefs over multiple objectives.
We show \algo{} achieves \sota{} performance across 8 \dl{} benchmarks in the multi-objective and single-objective setting.
Further, we demonstrate that \algo{} remains state-of-the-art at large scales by optimizing a vision-language model over three objectives.
\end{abstract}

\section{Introduction}\label{sec:intro}
Modern \dllong{} (\dl{}) pipelines \citep{vaswani-neurips17a, alphafold-nature21a, brown-neurips20a} are highly sensitive to the choice of their hyperparameters, the manual tuning of which has become an increasingly time-consuming and costly task. Despite substantial advances in algorithms for Hyperparameter Optimization (\hpo{})~\citep{bergstra-nips11a, li-iclr17a, falkner-icml18a, mallik-neurips23a}, many researchers continue to rely on manual tuning \citep{bouthillier-hal20a}, which allows intuitive incorporation of domain expertise and prior beliefs about the best performing \hp{} settings. 

While \hpo{} researchers have formulated desiderata for \hpo{} algorithms that include incorporating such user beliefs, existing research has focused exclusively on \soolong{}~\citep{ramachandran-kbs20a, souza-neurips20a, hvarfner-iclr22a, mallik-neurips23a}. However, for \dl{}, it is often necessary to optimize over several objectives, such as computational cost, training time, latency or fairness \citep{izquierdo-icml21a, schmucker-metalearn20a, salinas_multi-objective_2021, schneider_mo_interpret_tabular-2023}. Thus, integrating prior knowledge into \moolong{} is a crucial research area that remains unexplored.
We therefore adapt the desiderata for \hpo{} algorithms for \dl{} \citep{falkner-icml18a, mallik-neurips23a, franceschi_hyperparameter_2025} as follows:

\begin{table*}[htbp]
  \caption{
  Comparison of our algorithm \algo{} to prominent categories of multi-objective algorithms with respect to the identified desiderata. 
  The algorithmic categories include Evolutionary algorithms (EA, \eg{} \nsgaii{}, \smsemoa{}), \momflong{} algorithms (\momf{}, \eg{}, \moasha{}, \mohblong{}, \hbrwlong{}), and \molong{} Bayesian optimization (\mobo{}, \eg{}, \bo{} with random weights, \bo{} with \ehvi{}, \parego{}).
  A \cmark{} indicates that the method satisfies the criterion; a \textcolor{red}{\xmark}{} indicates it does not. (\cmark) denotes partial fulfillment or fulfillment with additional assumptions. }
  \label{table:desiderata}
  \centering
  \small
  \begin{tabular}{lccccccc}
    \toprule
    Criterion & \RS{} & EA & \momf{} &  \mobo{} & \algo{} \\
    \midrule
    Utilize cheap approximations   & \textcolor{red}{\xmark} & \textcolor{red}{\xmark}   & \cmark & \textcolor{red}{\xmark}   & \cmark \\
    Integrate multi-objective expert priors     & \textcolor{red}{\xmark} & \textcolor{red}{\xmark}   & \textcolor{red}{\xmark}  &  \textcolor{red}{\xmark}   & \cmark \\
    Strong anytime performance     & \textcolor{red}{\xmark} & \textcolor{red}{\xmark}   & \cmark & \textcolor{red}{\xmark}   & \cmark \\
    Strong final performance       & \textcolor{red}{\xmark} & (\cmark) & (\cmark) & \cmark     & \cmark \\
    \bottomrule
  \end{tabular}
\end{table*}

\begin{enumerate}
    

    \item \textbf{Utilize cheap approximations}: Modern \hpo{} algorithms must not only support optimization over multiple objectives, but should also be able to utilize cheap proxies of an objective function, if available, to speed up the optimization.

    \item \textbf{Integrate multi-objective expert priors}: Expert prior knowledge of hyperparameters is often available for real-world \dl{} tasks. A modern HPO algorithm must be able to utilize such beliefs over multiple objectives to speed up the optimization and be able to meaningfully recover from misleading prior information.
    \item \textbf{Strong anytime performance}: Multi-objective HPO algorithms must be compute-efficient, \ie{}, under limited budget, they must find candidates that significantly improve the dominated \hvlong{}.
    \item \textbf{Strong final performance}: 
    The ultimate goal of multi-objective \hpo{} is to find Pareto optimal candidates with strong dominated Hypervolume. As budgets grow larger, the algorithms should yield strong solutions.
\end{enumerate}


Table~\ref{table:desiderata} shows that existing \hpo{} algorithms satisfy at most half of the criteria. To address this gap, we propose \algo{}, which is the first \hpo{} algorithm to incorporate expert knowledge over the optima of multiple objectives and also leverages cheap approximations of expensive objective functions. Our \textbf{main contributions} are as follows:
\begin{itemize}
    \item We are the first to study expert priors for multiple objectives (\Secref{sec:ps}).
    \item We introduce \algo{}, a Bayesian optimization algorithm that integrates multi-objective expert priors in its acquisition function and exploits cheap proxy tasks in its \init{} (\Secref{sec:primo}).
    As such, \algo{} is the first HPO algorithm to meet all requirements of multi-objective HPO for DL (Table~\ref{table:desiderata}).
    \item We empirically demonstrate \sota{} performance of \algo{} across a variety of \dl{} benchmarks in the multi-objective \emph{and} single-objective setting, and show its robustness to different prior conditions (\Secref{exp:sota}). We also conduct several ablation studies (\Secref{subsec:ab_algos} and \appref{app:bo_ablation}) and a hyperparameter sensitivity analysis (Appendix \ref{app:bo_ablation}).
    \item We validate \algo{}'s applicability to large models by optimizing the vision-language model OpenCLIP \citep{radford-arxiv21a} over three objectives. Our case study shows that \algo{} maintains its superiority even when navigating competing objectives in a compute-intensive setting (\Secref{exp:realworld}).

\end{itemize}

\section{Multi-objective HPO with expert priors and cheap approximations}\label{sec:ps}

%

To capture all the above desiderata, we propose the novel problem formulation of optimizing a vector-valued objective function $f$, while exploiting cheap approximations of its individual objectives and expert priors. For background on HPO for DL, the utilization of cheap proxies, and the multi-objective case see \appref{app:more_bg}, and for a comparison to related problem formulations see \Secref{sec:relwork}.

\paragraph{Pareto optimality}

Typically, there does not exist a single best solution for multi-objective problems that minimizes all the objectives simultaneously. Rather, the goal is to find a strong set of solutions.

Given two candidates $\lambda_1$, $\lambda_2 \in \Lambda$,
we say that $\lambda_2$ dominates $\lambda_1$ \textit{if and only if} $f(\lambda_2) < f(\lambda_1)$.
A candidate $\lambda$ that is not dominated by any other candidate $\lambda'$ is called \textit{Pareto Optimal}, and the set of Pareto Optimal candidates is known as the \textit{Pareto Set} $\mathcal{P}$, defined as
\begin{equation}
    \mathcal{P} := \left\{\lambda \in \Lambda \,\middle|\, \nexists \lambda' \in \Lambda \text{ with } f(\lambda') < f(\lambda) \right\} \quad .
\end{equation}
\paragraph{Introducing multi-objective expert priors} To extend expert priors over a single objective \citep{hvarfner-iclr22a} to the multi-objective setting, we consider a factorized prior as follows. For each objective $f_i$ of the vector-valued function $f$, prior beliefs $\pi_{f_i}(\lambda)$ represent a probability distribution over the location of the optimum of $f_i$. Specifically, the prior will have a high value in regions that the user believes have an optimum. 
Formally, we define
\begin{equation}
    \pi_{f_i}(\lambda) = \mathbb{P} \left( f_i(\lambda) = \min_{\lambda' \in \Lambda} f_i(\lambda') \right) \, ,
\end{equation}
yielding the compound prior \(\Pi_f(\lambda) = \left\{ \pi_{f_i}(\lambda) \right\}_{i=1}^n \), i.e., the set of prior beliefs over the optima of the individual functions that comprise $f$.
For a discussion on our assumptions on priors see \Secref{sec:limits} and for details on the priors we consider in our experiments see \Secref{sec:exp_setup}. Further, we provide a discussion on the sources of prior knowledge in \appref{app_sec:prior_sources}.

\paragraph{Integrating multi-objective expert priors and cheap approximations} To also leverage cheap approximations of the individual objectives, let $\hat{f}_i(\lambda, z)$ denote the low-fidelity proxy for $f_i$, where hyperparameters $\lambda$ are evaluated at the fidelity level $z$, where $f_i(\lambda) = \hat{f}_i(\lambda, \maxfid)$. Therefore, our goal is to solve for $\mathcal{P}$ guided by $\Pi_{f}(\lambda)$
using inexpensive evaluations of $f$, while addressing the challenge that the priors may be misleading. 



\section{Related work}\label{sec:relwork}



Multi-objective optimization traditionally considers large budgets \citep{deb-book13a, deb_fast_2002, knowls-evoco06a, golovin_random_2020}; in \dl{}, however, budgets are constrained. 
Thus, to make HPO for \dl{} feasible, specialized strategies have been proposed. 
We discuss the strategies most related to ours here and provide a more expansive discussion in \appref{app:more_bg}.


\paragraph{User priors for single-objective optimization}
The integration of expert priors have been explored in a few works, but only for the \soolong{} case. Most similar to our approach (albeit for single-objective optimization) is \pb{} \citep{mallik-neurips23a}, which, in addition, to exploiting user priors also makes use of cheap approximations, in contrast to us, they use these throughout the optimization and not as an initial design. We explored adapting \pb{} to the multi-objective setting, but found it does not perform well (\Secref{sec:exp}). \pibo{} \citep{hvarfner-iclr22a}, like \algo{}, also augments the acquisition function with the priors, although it does so for a single objective only, can not utilize cheap approximations, and adapting it directly to the MO setting does not perform well under misleading priors (\Secref{sec:exp}).

\paragraph{Exploiting cheap approximations for multi-objective optimization} While exploiting expert priors is novel for multi-objective HPO, cheap proxies have been explored \citep{schmucker-metalearn20a, salinas_multi-objective_2021, schmucker-arxiv21a}. However, in contrast to our approach, not as an initial design. In our ablation study, we show that integrating cheap approximations as an initial design performs better overall (\Secref{subsec:ab_algos}).


\section{PriMO: prior informed multi-objective optimizer}\label{sec:method_primo}\label{sec:primo}

In this section, we introduce the first multi-objective HPO algorithm, \algo{} (Algorithm~\ref{algo:main_algo}), that leverages multi-objective user priors and fulfills all the desiderata of modern HPO.

We discuss how \algo{}, a Bayesian Optimization algorithm, makes use of multi-objective user priors via its acquisition function (Section \ref{sec:method_priors}) and cheap approximations of the objective functions with its initial design (Section \ref{sec:method_approx}). As \algo{} yields state-of-the-art for the multi-objective and single-objective setting (Section \ref{sec:exp}), we discuss how single-objective problems imply a special case of PriMO (Section \ref{sec:method_single_obj}). We provide additional details in \appref{app:algo_details}.

\subsection{Integrating multi-objective expert priors into Bayesian Optimization}\label{sec:method_priors}


We first choose one of the priors over multiple objectives uniformly at random during each iteration. We weight the acquisition function of \bo{} (Algorithm~\ref{algo:moprior_bo}) with the PDF of the selected prior, raised to an exponent \( \gamma = exp \left(- \nicefrac{n_{\bo{}}^2}{\mathit{n_d}} \right) \), where $n_{BO}$ and $n_d$ refer to the number of BO samples and the number of dimensions in the search space, respectively.
Unlike \pibo{}, where \(\gamma = \frac{10}{n}\), with $n$ referring to the n\textsuperscript{th} iteration, we reduce the overdependence on the prior by setting $\gamma$ to be inversely proportional to the square of the number of \bo{} samples. To formulate an acquisition function, we convert the vector-valued objective function into a \soolong{} problem, using a linear scalarization function~\citep{yoon_sequential_2009} with random weights, sampled at every optimization iteration, which is not only simple but also scalable with the number of objectives $n$:
\begin{equation}
    \min_{\lambda \in \Lambda}\;  \sum_{i=1}^{n} w_i\hat{f}_i(\lambda, \maxfid{}) \quad
    \text{s.t.} \quad  w_i \sim \mathcal{U}, \; w_i>0, \; \sum_{i=1}^{n}w_{i}=1 \, .
\end{equation}
Furthermore, to aid in recovery from misleading priors, we incorporate a simple exploration parameter \eps{}, which controls how often we augment the acquisition function with the prior. Thus, for \algo{}'s \eps{}-\bo{}, with priors over \textit{n} objectives, the acquisition function becomes
\begin{equation}\label{eq:acq}
\begin{split}
\alpha_{\epsilon\pi}(\lambda, \mathcal{D}) \triangleq 
\begin{cases}
&\alpha(\lambda, \mathcal{D}) \, , \quad \text{with prob. } \epsilon \\
&\alpha(\lambda, \mathcal{D}) \cdot \pi_{f_j}(\lambda)^{\exp\left(-\frac{n_{\bo{}}^2}{n_d}\right)} \, , \quad \text{with prob. } 1 - \epsilon,\; j \sim \mathcal{U} (1, \dots, n ) \, .
\end{cases}
\end{split}
\end{equation}

We use the $\texttt{qLogNoisyExpectedImprovement}$ acquisition function \citep{ament_unexpected_2023}, which we find to be the best overall compared to other prominent acquisition functions (see \appref{app:bo_ablation}). Furthermore, we set $\epsilon=0.25$ in our experiments, selecting the prior-weighted acquisition with a probability of 0.75, and perform a sensitivity study on \eps{} in \appref{app:eps_study}.

\subsection{An initial design to utilize cheap approximations}\label{sec:method_approx}

\begin{table}[t]
\centering
\caption{Comparison of the average HPO sampling times of \algo{} and \borw{}, and the evaluation cost of language modeling with a transformer on the 1B word benchmark.
}
\label{table:avg_hpo_time}
\begin{tabular}{lcc}
\toprule
Algorithm & Sampling time [s] & Model evaluation time [s] \\
\midrule
\algo{}   & 10 $\pm 1$ & 8307 $\pm 105$ \\
\borw{}   & 10 $\pm 1$ & 8741 $\pm 215 $\\
\bottomrule
\end{tabular}
\end{table}

To leverage cheap approximations of the objective functions, we propose an initial design strategy (Algorithm~\ref{algo:momf_init}) that exploits the strengths of multi-fidelity algorithms.
Specifically, we use a multi-fidelity algorithm in \algo{} to generate strong initial seed points at the maximum fidelity $\maxfid{}$ to speed up the optimization in the \bo{} phase afterward.

First, we set a threshold of (equivalent) full function evaluations based on the \init{} size. Once this threshold is reached, only maximum fidelity evaluations $\{ ( \lambda, \hat{f}(\lambda, \maxfid{} ) \}$ are included in the dataset $\mathcal{D}$ for use in \bo{}. Next, we choose one of the priors over multiple objectives uniformly at random during each iteration and the sampled initial points then aid the BO along with the decaying prior-augmented acquisition function (Equation \ref{eq:acq}).
We chose to use multi-objective asynchronous successive halving (\moasha{}) in our initial design due to its strong performance early on and since
as an infinite-horizon optimizer, it is budget invariant, resulting in a single continued optimization run without being restricted to discrete \shlong{} brackets. See Algorithm~\ref{algo:nondominated} and \ref{algo:mo-asha} for more details on \moasha{}.

\subsection{The single-objective setting}\label{sec:method_single_obj}

We also adapt \algo{} to the \solong{} setting as a special case of the original \molong{} design. The \init{} strategy of \algo{} replaces \moasha{} with the \asha{} scheduler. Instead of selecting a prior at random, as in the \molong{} case, we sample from the prior over the single objective. The \eps{}-greedy prior-augmented \bolong{} phase remains unchanged.

\subsection{Discussion}

\paragraph{Runtime Analysis}
We use scalarization to transform the multi-objective problem to a single-objective one in each BO step, therefore our runtime behaviour corresponds to classic BO. In our \eps{}-\bo{}, we use a Gaussian Processes which has the asymptotic time complexity of $\mathcal{O}(n^3)$.
We note that the runtime of a single \bo{} step is negligible in comparison to the model evaluation cost (for an example see \tabref{table:avg_hpo_time}). 
We further note that the runtime of \algo{}'s \init{} scales linearly with the number of objectives and quadratically with increasing number of candidate configurations in the higher rungs. However, that is still negligible compared to the evaluation cost of a \dl{} model.

\noindent
\begin{minipage}[tbp]{0.47\textwidth}
    \begin{algorithm}[H]
        \caption{Initial design strategy}
        \label{algo:momf_init}
        \footnotesize
        \begin{algorithmic}[1]
        \Function{\textnormal{\texttt{init}}}{$n_{\text{init}}, \Lambda, \eta, \minfid{}, \maxfid{}, f, \mathbf{w}$}
            \State $b \gets 0, \mathcal{D} \gets \emptyset$
            \While{$b < n_{\text{init}}$}
                \State $\lambda, z \gets \texttt{moasha}(\Lambda, \eta, \minfid{}, \maxfid{})$
                \State $\mathbf{y} \gets f(\lambda, z)$
                \If{$z = \maxfid{}$}
                    \State $\mathbf{y} \gets \mathbf{w}^\top \mathbf{y}$
                    \State $\mathcal{D} \gets \mathcal{D} \cup \{(\lambda, \mathbf{y})\}$
                \EndIf
                \State $b \gets b + \frac{z}{\maxfid{}}$
            \EndWhile
            \State \Return $\mathcal{D}$
        \EndFunction
        \end{algorithmic}
    \end{algorithm}
\end{minipage}%
\hfill
\begin{minipage}[tbp]{0.48\textwidth}
    \begin{algorithm}[H]
        \caption{BO step with 
        multi-obj.\
        priors}
        \label{algo:moprior_bo}
        \footnotesize
        \begin{algorithmic}[1]
            \Function{\textnormal{\texttt{moprior\_bo}}}{$\Lambda, \mathcal{D}, \Pi_{f}, n_{\text{BO}} , \epsilon$}
                \State Select prior $\pi_{f_j}$, where $j \sim \mathcal{U}(1, \dots, n)$
                \State $\gamma \gets \exp(-n_{\text{BO}}^2 / n_d)$
                \State $u \sim \mathcal{U}(0,1)$
                \If{$u < \epsilon$}
                    \State $\tilde{\alpha}(\lambda) := \alpha(\lambda, \mathcal{D})$
                \Else
                    \State $\tilde{\alpha}(\lambda) := \alpha(\lambda, \mathcal{D}) \cdot \pi_{f_j}(\lambda)^{\gamma}$
                \EndIf
                \State $\lambda \gets \arg\max_{\lambda \in \Lambda} \tilde{\alpha}(\lambda)$
                \State \Return $\lambda$
            \EndFunction
        \end{algorithmic}
    \end{algorithm}
\end{minipage}
\begin{algorithm}[!ht]
    \caption{\algo{}}
    \label{algo:main_algo}
    \footnotesize
    \begin{algorithmic}[1]
        \State \textbf{Input:} Objective $f$, search space $\Lambda$ with dimension $n_d$, priors $\Pi_{f} = \{ \pi_{f_i}(\lambda) \}_{i=1}^n$, \init{} size $n_{\text{init}}$, reduction factor $\eta$, fidelity range $[\minfid{}, \maxfid{}]$, budget $B$ and exploration parameter \eps{}.
        \Function{\textnormal{\algo{}}}{Input}
                \State $\mathcal{D} \gets \texttt{init}(n_{\text{init}}, \Lambda, \eta, \minfid{}, \maxfid{}, f, \mathbf{w})$
            \State  $b \gets n_{\text{init}}$, $n_{\text{BO}} \gets 0$
            \While{$b < B$}
                \State $\bm{\lambda}_{new} \gets \texttt{moprior\_bo{}}(\Lambda, \mathcal{D}, \Pi_{f}, n_{\text{BO}}, \epsilon)$
                \State $n_{\text{BO}} \gets n_{\text{BO}} + 1$
                \State $\mathbf{y} \gets f(\bm{\lambda}_{new}, \maxfid{})$
                \State Sample weights $\mathbf{w} \sim \mathcal{U}(0,1)^n$ and normalize
                \State $ \mathbf{y} \gets \mathbf{w}^\top \mathbf{y}$
                \State $\mathcal{D} \gets \mathcal{D} \cup \{(\bm{\lambda}_{new}, \mathbf{y})\}$
                \State $b \gets b + 1$
            \EndWhile
            \State \Return $\mathcal{P}_f(\mathcal{D})$
        \EndFunction
    \end{algorithmic}
\end{algorithm}

\begin{figure}[H]
    \centering
    \includegraphics[width=0.5\linewidth]{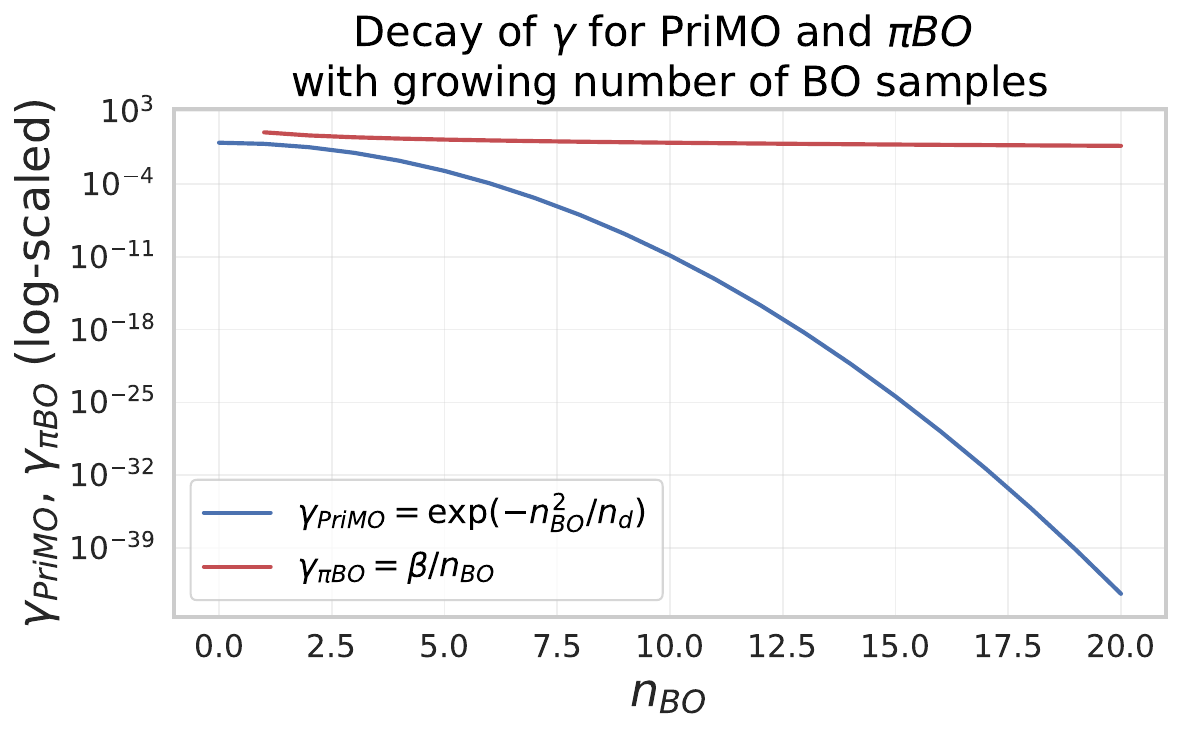}
    \caption{Decay of the prior exponent $\gamma$ with the increase in the number of \bo{} samples for \algo{} and the original \pibo{} implementation. Here, $n_d$ refers to the search spaces dimensions, which we fix to 4 - the size of the search space of the \texttt{PD1} benchmarks. We note that the \pibo{} authors set $\beta=10$.}
    \label{fig:gamma_decay}
\end{figure}

\paragraph{Behavior under highly correlated priors}
When the priors for two objectives are strongly positively correlated, the overall dependence on their quality increases. If the priors are helpful for both objectives, \algo{}'s performance will improve, while the converse is true if the priors for both objectives are misleading. However, under mixed prior conditions, their effects may cancel out or the prior over one objective may suppress the effect of the other depending on the strength of the correlation between them.

\paragraph{$\gamma$ decay comparison}
In \Figref{fig:gamma_decay}, we plot the decay of $\gamma$, the exponent of the prior in the weighted acquisition function, in case of \algo{} and the original \pibo{} implementation, with increasing number of \bolong{} samples. We show that the $\gamma$, and hence the strength of the prior, decays much faster in \algo{} in comparison to \pibo{}, which helps to recover from misleading prior information much quicker.

\section{Experiments}\label{sec:exp}

To empirically demonstrate that \algo{} fulfills the desiderata outlined in the \Secref{sec:intro}, we address the following research questions.

\begin{itemize}
    \item[\rqa{}:] Does \algo{} achieve state-of-the-art performance for multi-objective optimization?
    \item[\rqb{}:] Does \algo{} maintain \sota{} performance in the \solong{} setting?
    \item[\rqc{}:] Can \algo{} effectively leverage \molong{} expert priors, as well as recover from misleading priors, to maintain its robustness?
    \item[\rqd{}:] Does \algo{} maintain its superior performance when scaling to a compute-intensive application with three objectives?
    \item[\rqe{}:] How does the contribution of the proposed \eps{}-\bo{} compare to that of the \init{}? 
    \item[\rqf{}:] Are the choices of scalarization and acquisition function for \algo{} justified?
    
\end{itemize}


After providing details on our experimental setup and baselines (Section \ref{sec:exp_setup}), we first show \algo{}'s state-of-the-art performance in the multi-objective and single-objective setting across our 8 \dllong{} benchmarks in Section \ref{exp:sota} (answering \rqa{}, \rqb{} and the first part of \rqc{}). We then discuss its robustness across all prior conditions (answering the second part of \rqc{}), and demonstrate how \algo{} scales to real world applications and problems with more than 2 objectives in \Secref{exp:realworld} (\rqd{}). Finally, we provide ablation studies in Section \ref{subsec:ab_algos} (answering \rqe{} and \rqf{}). 
In \appref{app:all_exp} we provide detailed case-level analysis based on Pareto fronts, and hypervolume plots under all prior combinations.


\subsection{Experimental setup}\label{sec:exp_setup}



\paragraph{Evaluation protocol} We base our multi-objective evaluation on the mean dominated hypervolume across 25 seeds and report relative rankings of the algorithms across budgets. Each optimizer-benchmark-seed combination was run for \nfevals{} equivalent full function evaluations, corresponding to typical budgets in practical Deep Learning. We give more details on our evaluation protocol in \appref{app_sec:study_setup}, provide additional experiments and analysis in \appref{app:all_exp}, and conduct a statistical significance analysis in \appref{app:sig_analysis}.

\paragraph{Benchmarks} For our multi-objective experiments, we use 8 benchmarks representing image classification, language translation and learning curves for Deep Neural Networks. We chose 4 \lcbench{} benchmarks from the \yahpogym{} Suite \citep{pfisterer_yahpo_2022} and 4 from the \pdone{} \citep{wang_pre-trained_2024} set of benchmarks. We select the corresponding validation error and training cost metrics as objectives. 
For the single-objective setting, we stick to the benchmarks used by \citet{mallik-neurips23a}. Note that for our main benchmark suites - \pdone{} and \yahpogym{}, we have 4 and 7 dimensional search spaces respectively, with 2 objectives for each, which are the most common settings for practical \dl{}. 



\paragraph{Baselines} We give an overview of all our baselines here and provide additional details in Appendix \ref{app_sec:study_opts}. We compare \algo{} against a host of prominent \mo{} baselines representing different classes of optimization algorithms for \mo{}, which include model-based \momf{} (\momfbo{}) \citep{irshad_leveraging_2024}, scalarized \bolong{} approaches like \bo{} with random weights (\borw{}) and \parego{} \citep{knowls-evoco06a}, \bo{} with \ehvilong{} acquisition function (qLogNEHVI) \citep{ament_unexpected_2023}, \mflong{} optimizers such as \hbrwlong{} (\hbrw{}) \citep{schmucker-metalearn20a} and multi-objective asynchronous successive halving (\moasha{}) \citep{schmucker-arxiv21a}, and an evolutionary algorithm - \nsgaii{} \citep{deb_fast_2002}.
While no multi-objective in the literature leverage expert priors, we augment single-objective approaches to provide strong prior-based baselines and refer to Appendix \ref{app:prior_baselines} for a comparison.
For our experiments in the single-objective setting, we compare \algo{} against \bo{} and \hblong{} \citep{li-iclr17a}, and existing single-objective algorithms that can leverage expert priors, i.e., \pb{}-\bo{} \citep{mallik-neurips23a} and \pibo{} \citep{hvarfner-iclr22a}.

\paragraph{Priors} We study the effect of different prior conditions: both objectives have good priors, both objectives have bad priors, the average over mixed good and bad prior combinations, as well as the average over all combinations. For generating priors, we adapt the protocols in the literature on single-objective optimization \citep{hvarfner-iclr22a, mallik-neurips23a}, for details see Appendix \ref{app_sec:study_priors}.

\subsection{PriMO's state-of-the-art performance and robustness to prior conditions}\label{exp:sota}


\begin{figure*}[tb]
    \centering
    \includegraphics[width=1.0\linewidth]{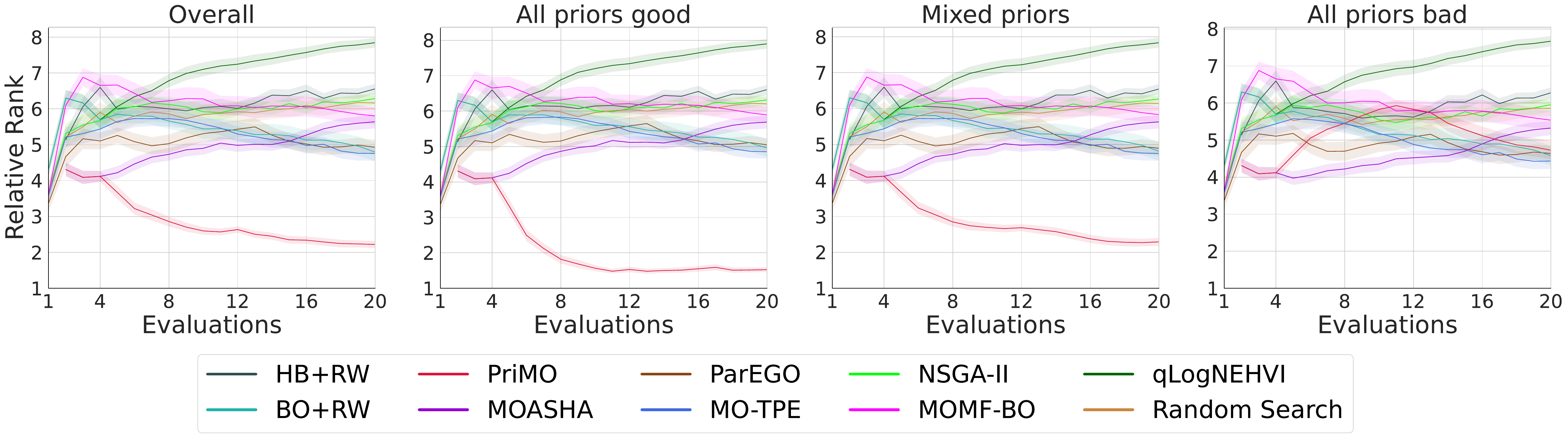}
    \caption{Mean relative ranks $\pm$ 1 standard error of \algo{} and prominent multi-objective algorithms across benchmarks and seeds under various prior conditions.}
    \label{fig:ovrank_subset1_all_4}
\end{figure*}
\begin{figure}[t]
    \centering
    \includegraphics[width=0.7\linewidth]{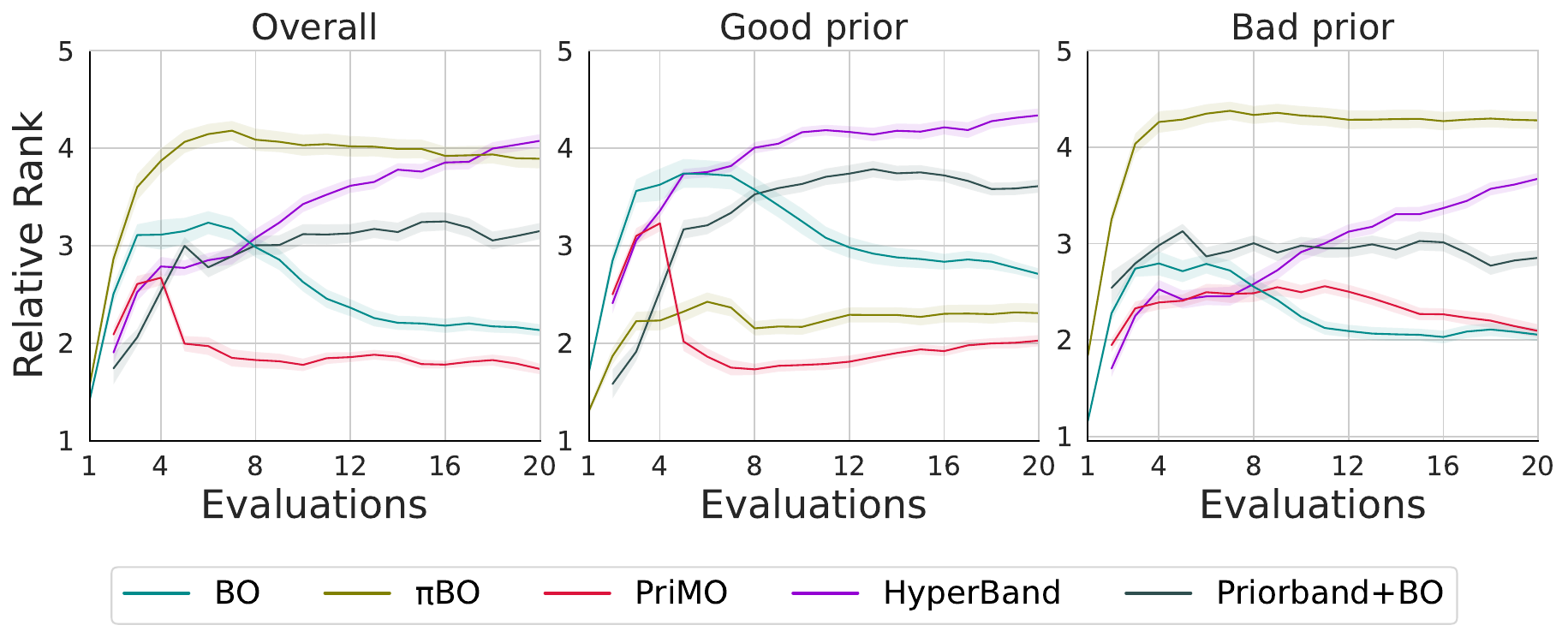}
    \caption{Mean relative ranks $\pm$ 1 standard error in the single-objective setting across benchmarks and seeds under various prior conditions.
    }
    \label{fig:ovrank_so_all_3}
\end{figure}

\paragraph{State-of-the art in the multi-objective setting}

\Figref{fig:ovrank_subset1_all_4} demonstrates that, overall, \algo{} maintains the strongest anytime performance and achieves the best final performance in terms of relative rankings across all benchmarks (\rqa{}). Under good priors the relative ranking gap between \algo{} and the second best optimizer, \motpe{}, is even more pronounced. 
\algo{} shows a strong start under good priors, and is the best performing algorithm early on. We attribute this to the \init{} which provides a strong head-start before the \bo{} phase. Using the configurations sampled by the \init{}, the \bo{} phase of \algo{} maintains its strong anytime performance, and is able to effectively utilize the priors, achieving \sota{} final performance across all benchmarks (\rqc{}).
Designed with the practical \dl{} use case in mind, where most practitioners operate on modest budgets, \algo{} reduces its dependence on priors after approximately 10 \bo{} samples, governed by our chosen $\gamma$ setting (see \Secref{sec:primo}). 

\paragraph{State-of-the art in the single-objective setting}
\Figref{fig:ovrank_so_all_3} shows \algo{}'s \sota{} performance in the \solong{} setting. \algo{} is overall the best choice for \solong{} \hpo{} demonstrating the strongest anytime and final performance (\rqb{}). Under good priors, \pb{} and \pibo{} show stronger starts as is expected with their prior-based initial sampling strategies, but are quickly outperformed by \algo{} within a few full function evaluations.

\paragraph{PriMO is robust to bad priors}
Even when the priors for all objectives are bad (\Figref{fig:ovrank_subset1_all_4}, right) \algo{} shows a remarkable recovery after a poor initial performance and by the end of the optimization budget, nearly matches \motpe{}, \parego{} and \borw{} (with no statistically significant difference in performance - \appref{app:sig_analysis}), resulting in a competitive final performance (\rqc{}). Mixed and \ova{} prior conditions show similar trends where \algo{} is the best performing algorithm early on and ranks significantly better than all other baselines by the final iteration.

\begin{figure}[t]
    \centering
    \includegraphics[width=1\linewidth]{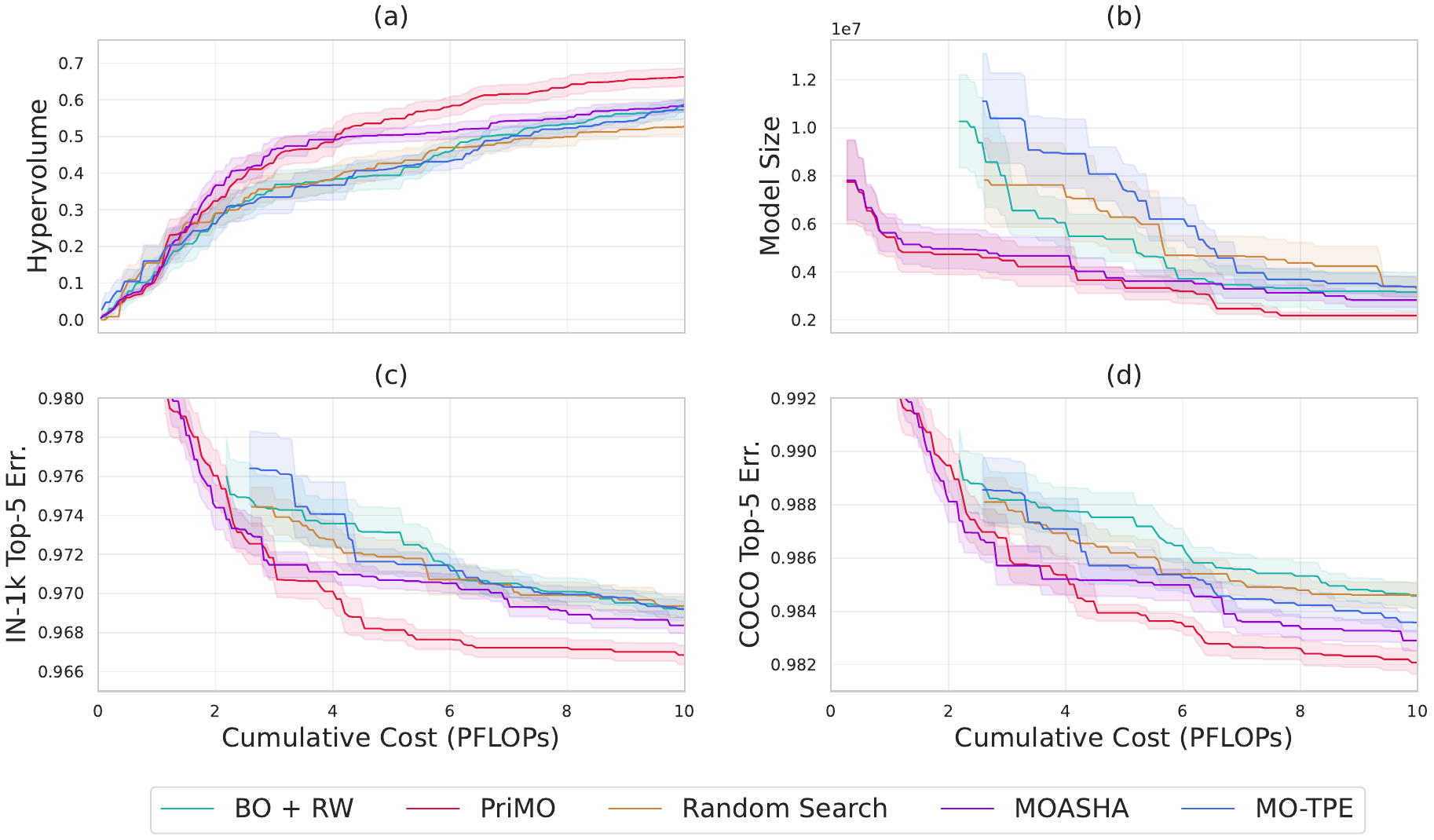}
    \caption{Case study on the OpenCLIP pipeline across 3 objectives over 20 seeds. [Panel (a)] Mean dominated hypervolume vs. cumulative compute cost (in PFLOPs). [Panels (b), (c), and (d)] The trajectories of the best-found (incumbent) values for model parameter size, ImageNet1k error, and MS-COCO error. \algo{} outperforms all other baselines, achieving both the highest hypervolume and the best individual values for each objective within the 10 PFLOP budget.}
    \label{fig:vlm}
\end{figure}

\subsection{Large-scale case study}\label{exp:realworld}
Beyond surrogate benchmarks, we evaluate \algo{} on a real-world pipeline demanding a substantial compute budget: pretraining OpenCLIP across 3 objectives comprising model size and the error rates (1 - Acc@5) on two downstream tasks (image classification on ImageNet1k and a retrieval task on MS-COCO).
In this case study, we do not study different prior strengths in a controlled setting, but formulate expert priors based on prior experience with the OpenCLIP pipeline.
Due to the computational costs involved, we compare \algo{} only against the strongest baselines of the previous experiments (\moasha{}, \borw{}, and \motpe{}), and \RS{}. As shown in \Figref{fig:vlm}, given a cumulative budget of 10 PFLOPs and aggregating over 20 seeds, \algo{} achieves a significantly higher dominated hypervolume and superior convergence across the objective space (\rqd{}). Further details on the case study can be found in \appref{app_sec:real_world}.

\subsection{The significant contribution of the \eps{}-\bo{} over the \init{} strategy}\label{subsec:ab_algos}
\begin{figure*}[t]
    \centering
    \includegraphics[width=1.0\linewidth]{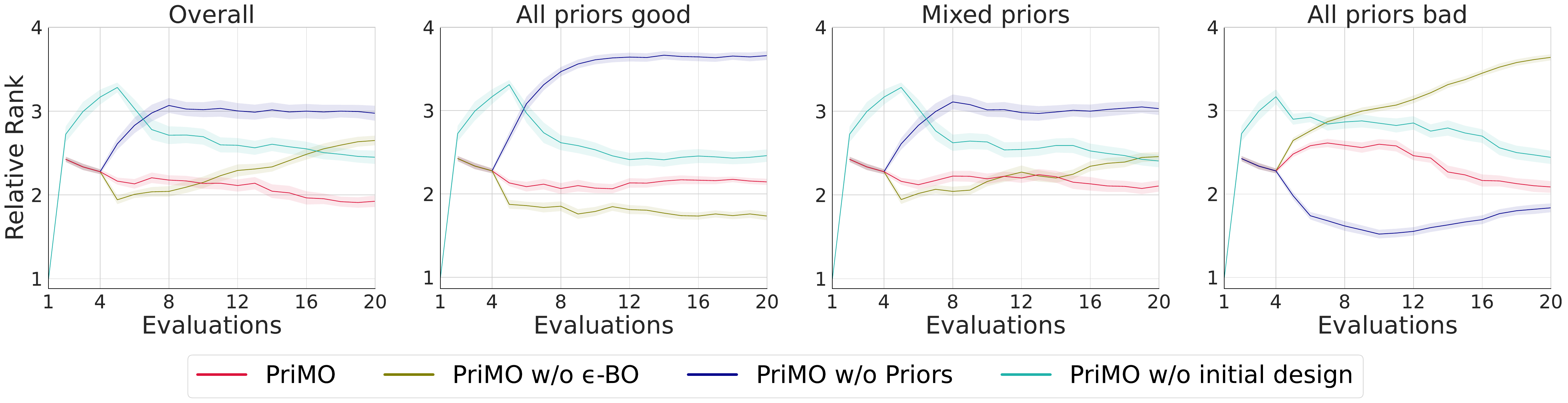}
    \caption{Mean relative ranks $\pm$ 1 standard error of various ablations of \algo{} across benchmarks and seeds under various prior conditions.}
    \label{fig:ovranks_ablations_algos_all}
\end{figure*}

We consider different design ablations of \algo{} in \Figref{fig:ovranks_ablations_algos_all} to answer \rqe{} and, although overall, find that all components of \algo{} are important, the contribution of the \eps{}-\bo{} is rather significant.
We divide \algo{} into its constituent components, namely - the \init{}, MO-Priors and the \eps{}-\bo{} (including the random weights), and label each design ablation with respect to the component(s) that were removed. We note that \algo{} without Priors and \algo{} without \eps{}-\bo{} correspond to vanilla \bo{} and \piborw{} respectively, both augmented with our \init{} strategy.

We find that the \init{} strategy gives a substantial early boost, as all ablations of \algo{} using the \init{} start off much stronger than \algo{} without the \init{} (\rqe{}). This initial advantage tends to persist for most ablations until the \bo{} phase for \algo{} without the \init{}. Overall, while the \momf{} \init{} provides meaningful early speedups to \bo{}, it does not sustain strong performance in the long run unless paired with the \eps{}-\bo{}.

\algo{} without Priors and \algo{} without \eps{}-\bo{} are two of the worst performing ablations overall, highlighting the importance of a prior-based \bo{} design, coupled with the \eps{}-greedy optimization strategy in \eps{}-\bo{}.
These findings, taken together, support our final design choice for \algo{} (\rqe{}).

In \appref{app:bo_ablation}, we perform another ablation study on the scalarization method and acquisition function used in \algo{}, to answer \rqf{}.

\section{Limitations and discussion}\label{sec:limits}
Despite its impressive empirical performance and our extensive ablation study and sensitivity analysis to \algo{}'s hyperparameters, we do not provide theoretical guarantees for \algo{}. We do experiment with Hypervolume scalarization \citep{golovin_random_2020} which has provable guarantees to converge to non-convex Pareto fronts, however, we see that \algo{} with the default Linear scalarization outperforms \algo{} with Hypervolume Scalarization (\appref{app:bo_ablation}) and opt for strong empirical performance over theoretical guarantees.

In line with previous work \citep{souza-neurips20a, hvarfner-iclr22a, mallik-neurips23a}, we consider Gaussian distributions for our priors, where the mean corresponds to a configuration that is expected to perform well, although \algo{} supports priors with any distribution. While it may be more beneficial to formulate priors based on an approximate Pareto front, \algo{} already achieves \sota{} performance using factorized priors.

We chose our benchmarks to represent practically relevant DL settings; naturally, most of our benchmarks have low-dimensional hyperparameter spaces and two objectives that correspond to some performance metric and a compute cost metric. While benchmarks with more than three objectives are known to have highly correlating objectives \citep{li2021hwnasbench, bansal-jahs-neurips22a, dou2023eahasbench, NEURIPS2024_6ffdbf06}, we analyze \algo{}'s runtime for more than two objectives and investigate its empirical performance for three objectives in our case study.

\section{Conclusion}\label{sec:conc}

\algo{} distinguishes itself as the first algorithm to integrate multi-objective expert priors, leading to state-of-the-art performance. As such, \algo{} is, to date, the only \hpo{} algorithm that fulfills all the desiderata of modern \hpo{}, making it fit for efficient optimization under constrained budgets for practical Deep Learning.

\subsection*{AI use statement}

In this work, we have only used generative AI tools to aid with the visualization of the results, and not for any other tasks including the code or writing of the manuscript. The new algorithm proposed in this paper is an original contribution of the authors and we take responsibility for the full final content of this work.

\subsection*{Reproducibility statement}

To ensure reproducibility, we adhere to and include a reproducibility checklist in Appendix~\ref{app:checklist}. In our justifications for the checks, we link to all relevant sections, appendices, and supplementary materials. We considered the checklists in use by NeurIPS and AutoML-Conf, but chose the latter as it is more comprehensive.





\subsubsection*{Acknowledgments}
Frank Hutter acknowledges the financial support of the Hector Foundation. The authors acknowledge support by the state of Baden-Württemberg through bwHPC. We acknowledge funding by the European Union (via ERC Consolidator Grant DeepLearning 2.0, grant no.~101045765). Views and opinions expressed are however those of the author(s) only and do not necessarily reflect those of the European Union or the European Research Council. Neither the European Union nor the granting authority can be held responsible for them. \begin{center}\includegraphics[width=0.3\textwidth]{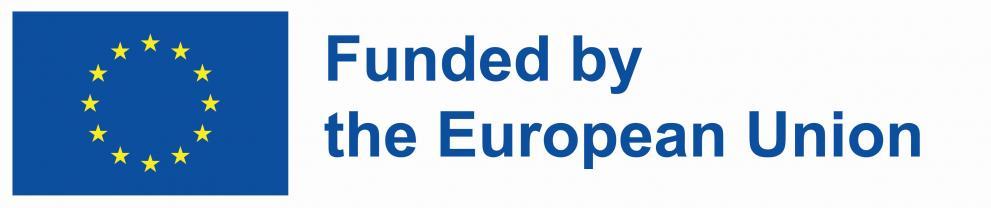}\end{center}



\bibliographystyle{iclr2027_conference}
\bibliography{bib/strings,bib/lib,bib/proc, bib/MO, bib/MO_MF, bib/General_HPO_and_Benchmarking, bib/Software, bib/other}


\newpage
\appendix

\section{Reproducibility Checklist}\label{app:checklist}
\begin{enumerate}
\item For all authors\dots
  \begin{enumerate}
  \item Do the main claims made in the abstract and introduction accurately
    reflect the paper's contributions and scope?
    \answerYes{Yes, in our list of contributions in the introduction we reference the corresponding sections for each contribution we claim..}
  \item Did you describe the limitations of your work?
    \answerYes{See \Secref{sec:limits}.}
  \item Did you discuss any potential negative societal impacts of your work?
    \answerNA{}
  \item Did you read the ethics review guidelines and ensure that your paper
    conforms to them? (see \url{https://2026.automl.cc/ethics-and-accessibility-guidelines/})
    \answerYes{}
  \end{enumerate}
\item If you ran experiments\dots
  \begin{enumerate}
  \item Did you use the same evaluation protocol for all methods being compared (e.g.,
    same benchmarks, data (sub)sets, available resources, etc.)?
    \answerYes{We describe our evaluation protocol in \Secref{sec:exp_setup}.}
  \item Did you specify all the necessary details of your evaluation (e.g., data splits,
    pre-processing, search spaces, hyperparameter tuning details and results, etc.)?
    \answerYes{See Appendix~\ref{app_sec:study_priors}, \ref{app_sec:study_opts}, \ref{app_sec:study_benches}, and \ref{app_sec:study_setup}.}
  \item Did you repeat your experiments (e.g., across multiple random seeds or
    splits) to account for the impact of randomness in your methods or data?
    \answerYes{See \Secref{sec:exp_setup}.}
  \item Did you report the uncertainty of your results (e.g., the standard error
    across random seeds or splits)?
    \answerYes{See \Secref{sec:exp} and \appref{app:all_exp}.}
  \item Did you report the statistical significance of your results?
    \answerYes{See \appref{app:sig_analysis}.}
  \item Did you use enough repetitions, datasets, and/or benchmarks to support
    your claims?
    \answerYes{}
  \item Did you compare performance over time and describe how you selected the
    maximum runtime?
    \answerYes{We compare across budgets and provide a description in \Secref{sec:exp_setup}.}
  \item Did you include the total amount of compute and the type of resources
    used (e.g., type of \textsc{gpu}s, internal cluster, or cloud provider)?
    \answerYes{See \appref{app:resources}.}
  \item Did you run ablation studies to assess the impact of different
    components of your approach?
    \answerYes{See \Secref{subsec:ab_algos}.}
  \end{enumerate}
\item With respect to the code used to obtain your results\dots
  \begin{enumerate}
\item Did you include the code, data, and instructions needed to reproduce the
    main experimental results, including all dependencies (e.g.,
    \texttt{requirements.txt} with explicit versions), random seeds, an instructive
    \texttt{README} with installation instructions, and execution commands
    (either in the supplemental material or as a \textsc{url})?
    \answerYes{See \appref{app_sec:repo}.}
  \item Did you include a minimal example to replicate results on a small subset
    of the experiments or on toy data?
    \answerYes{In the experiment repository provided in \appref{app_sec:repo}.}
  \item Did you ensure sufficient code quality and documentation so that someone else
    can execute and understand your code?
    \answerYes{}
  \item Did you include the raw results of running your experiments with the given
    code, data, and instructions?
    \answerYes{See \appref{app_sec:repo}.}
  \item Did you include the code, additional data, and instructions needed to generate
    the figures and tables in your paper based on the raw results?
    \answerYes{See \appref{app_sec:repo}.}
  \end{enumerate}
\item If you used existing assets (e.g., code, data, models)\dots
  \begin{enumerate}
  \item Did you cite the creators of used assets?
    \answerYes{}
  \item Did you discuss whether and how consent was obtained from people whose
    data you're using/curating if the license requires it?
    \answerYes{See \appref{app:licenses}.}
  \item Did you discuss whether the data you are using/curating contains
    personally identifiable information or offensive content?
    \answerNA{}
  \end{enumerate}
\item If you created/released new assets (e.g., code, data, models)\dots
  \begin{enumerate}
    \item Did you mention the license of the new assets (e.g., as part of your
    code submission)?
    \answerYes{See \appref{app:licenses}.}
    \item Did you include the new assets either in the supplemental material or as
    a \textsc{url} (to, e.g., GitHub or Hugging Face)?
    \answerYes{See \appref{app_sec:repo}.}
  \end{enumerate}
\item If you used crowdsourcing or conducted research with human subjects\dots
  \begin{enumerate}
  \item Did you include the full text of instructions given to participants and
    screenshots, if applicable?
    \answerNA{}
  \item Did you describe any potential participant risks, with links to
    institutional review board (\textsc{irb}) approvals, if applicable?
    \answerNA{}
  \item Did you include the estimated hourly wage paid to participants and the
    total amount spent on participant compensation?
    \answerNA{}
  \end{enumerate}
\item If you included theoretical results\dots
  \begin{enumerate}
  \item Did you state the full set of assumptions of all theoretical results?
    \answerNA{}
  \item Did you include complete proofs of all theoretical results?
    \answerNA{}
  \end{enumerate}
\end{enumerate}

\section{Background and additional related work}\label{app:more_bg}

\subsection{Hyperparameter optimization for Deep Learning}\label{sec:hpo4dl}

\paragraph{Multi-fidelity optimization}\label{sec:ps_mfo}

The high computational cost of \dl{} model evaluations has  motivated research in \mfolong{}. Multi-fidelity (\mf{}) \citep{kandasamy-icml17a} optimizers use \textit{cheap proxies} to approximate promising candidates and speed up the search. Bandit-based methods \citep{jamieson-aistats16a, li-iclr17a} are the most popular in the Automated Machine Learning community for \mfolong{}. These have been further extended by replacing their \RSlong{} (\RS{}) component with evolutionary \citep{awad-ijcai21a} and model-based \citep{falkner-icml18a} search, and increasing efficiency for large-scale parallelization \citep{li-mlsys20a}.

Instead of optimizing the expensive objective function $f$ as a blackbox, \mfolong{} leverages evaluations of $f$ at lower fidelities. For example, when training a Neural network with a particular \hp{} configuration for 100 epochs, a lower-fidelity proxy would be the validation score obtained by training the model with the same \hp{} configuration for 15 epochs.
More formally, for a \hp{} configuration $\lambda \in \Lambda$ at a fidelity level $z \in Z$ where $ Z \coloneqq \{\minfid{},...,\maxfid{} \}, |Z| = m$ is the fidelity space, a cheap proxy function of $f$ is defined as $\hat{f}(\lambda, z)$. Therefore, when $z = \maxfid{}$ (the maximum fidelity), the proxy function $\hat{f}$ converges to the true objective function $f$. Hence, $f = \hat{f}(\lambda, \maxfid{})$.

In an optimization setup with \textit{continuations}, the function evaluation $\hat{f}(\lambda, z)$ for a configuration $\lambda$ at fidelity $z$ can be continued up to a fidelity $z'$ to yield $\hat{f}(\lambda, z')$, given $z < z'$. For example, let us assume that we would like to train a network with a \hp{} configuration $\lambda$ for a total of 200 epochs, and have already trained it with $\lambda$ for 50 epochs. Then we can simply continue training with $\lambda$ for 150 more epochs instead of restarting from scratch. For such a continual setup, we define \textit{equivalent function evaluations} as $z/\maxfid{}$.

\paragraph{Prior-based optimization for a single objective}

Prior-based single objective optimization can be defined as solving
\begin{equation}
    \arg\min_{\lambda \in \Lambda} f(\lambda), \quad \text{guided by } \pi(\lambda),
\end{equation}
where prior $\pi(\lambda)$ is a probability distribution over the location of the optimum of the objective function $f$.

PrBO \citep{souza-neurips20a} combines expert prior distributions $P_g(\lambda)$ and $P_b(\lambda)$ with respective models $M_g(\lambda)$ and $M_b(\lambda)$ in a Tree-structured Parzen Estimator \citep{bergstra-nips11a} (TPE)-based approach to construct \textit{pseudo-posteriors} $g(\lambda)$ and $l(\lambda)$ respectively. The candidates are then chosen from these pseudo-posteriors by maximizing the \ei{} as described in \citet{bergstra-nips11a}.
\pibo{} \citep{hvarfner-iclr22a} directly augments the acquisition function $\alpha$ with the unnormalized user-specified prior distribution $\pi(\lambda)$ which decays over time, controlled by a parameter $\beta$: \( \alpha^{n}_{\pi}(\lambda) = \alpha(\lambda) \cdot \pi(\lambda)^\frac{\beta}{n} \), where $n$ refers to the $n^{th}$ iteration. Unlike PrBO, \pibo{} generalizes to acquisition functions other than EI and offers convergence guarantees. However, as we saw see in \Secref{sec:exp}, \pibo{}'s longer dependence on the Priors has major downsides. \algo{} addresses this issue using a novel \mo{}-priors-based augmentation of the \bo{} component that we introduced in \Secref{sec:primo}.

\citet{mallik-neurips23a} introduce \pb{} which extends the integration of expert priors to \mfolong{}. \pb{} uses a novel \textit{ensemble sampling policy (ESP)} $\mathcal{E}_{\pi}$, which combines random sampling $\mathcal{U}(\cdot)$, prior-based sampling $\pi(\cdot)$ and incumbent-based sampling $\hat{\lambda}(\cdot)$, with their proportions denoted by $p_{\mathcal{U}}$, $p_{\pi}$ and $p_{\hat{\lambda}}$ respectively. Initially, $\hat{\lambda}(\cdot)$ is inactive. Given the constraint $p_{\mathcal{U}} + p_{\pi} = 1$, $\mathcal{E}_{\pi}$ selects from $\mathcal{U}(\cdot)$ and $\pi(\cdot)$ according to $p_{\mathcal{U}}$ and $p_{\pi}$. When $\hat{\lambda}(\cdot)$ becomes active, $p_{\pi}$ is split into $p_{\pi}$ and $p_{\hat{\lambda}}$ according to weighted scores $\mathcal{S}_\pi$ and $\mathcal{S}_{\hat{\bm{\lambda}}}$, calculated by first computing the likelihood of the top performing configurations under $\pi(\cdot)$ and $\hat{\lambda}(\cdot)$, which capture how much \textit{trust} should be placed on each.

While the aforementioned algorithms efficiently integrate user priors in the \hpo{} problem, they only apply to the \soolong{} case. To the best of our knowledge, we are the first to incorporate priors over multiple objectives, whilst also employing a novel \init{} strategy to leverage cheap proxies of the objective function.

\subsection{Multi-objective optimization}\label{app:bg_mo}

For many real-world problems we are often interested in optimizing not one, but multiple, potentially competing objectives. \mo{} \citep{srinivas_muiltiobjective_1994, deb_fast_2002, knowls-evoco06a, zhang-moead-2007} deals with optimizing a \textit{vector-valued objective function} \textbf{$f(\lambda)$} composed of $n$ distinct objective functions, where $f: \lambda \rightarrow \mathbb{R}^n$, $\lambda \in \mathbb{R}$. Without loss of generality, we assume minimization of all objectives.





\paragraph{More on Pareto Optimality}

As mentioned earlier in \Secref{sec:ps}, given two candidates $\lambda_1$, $\lambda_2 \in \Lambda$, we say that $\lambda_2$ dominates $\lambda_1$ \textit{if and only if} $f(\lambda_2) < f(\lambda_1)$. Formally, we write $\lambda_2$ $\prec$ $\lambda_1$.
Additionally, for $f(\lambda_2) \leq f(\lambda_1)$, we write $\lambda_2$ $\preceq$ $\lambda_1$ and say that $\lambda_1$ weakly dominates $\lambda_2$.

For a vector-valued function \textit{f}, we say that $\lambda_2$ Pareto dominates $\lambda_1$, i.e. $\lambda_2$ $\prec$ $\lambda_1$ under two conditions:
\begin{itemize}
    \item $\forall i \in \{1,..., n\}: f_i(\lambda_2) \leq f_i(\lambda_1)$, and,
    \item $\exists k \in \{1,..., n\}: f_k(\lambda_2) < f_k(\lambda_1)$ .
\end{itemize}

\paragraph{Pareto Front}

The set of solutions, i.e., set the corresponding values of an \mo{} function for each of the Pareto Optimal candidates is called the \textit{Pareto Front}. Formally, a Pareto front is defined as:
\begin{equation}
    \mathcal{F} = 
    \left\{ f(\lambda) \in \mathbb{R}^n 
    \,\middle|\, \lambda \in \Lambda, \nexists \lambda' \in \Lambda 
    \text{ with } f(\lambda') < f(\lambda) \right\} \quad .
\end{equation}

\paragraph{Hypervolume indicator}
The true Pareto front of a real-world \mo{} problem is generally unknown. Thus, the goal of \mo{} Optimization algorithms is to return a set of non-dominated candidates from which we can obtain an \textit{approximated Pareto front}. To assess the quality of this approximation, the \textit{S-Metric} or \textit{Hypervolume (\hv{}) Indicator} \citep{zitzler_multiobjective_1998} is the most frequently used measure as it does not require prior knowledge of the true Pareto front.

Given a reference point $r$ and an approximate Pareto set $\mathcal{A}$, the Hypervolume Indicator $\mathcal{H}$ is defined as:
\begin{equation}
    \mathcal{H}_r(\mathcal{A}) = 
    \mu \left( 
    \{x \in \mathbb{R}^n 
    | \exists a \in A : a \leq x \cap x \leq r \} 
    \right) \quad ,
\end{equation}

where $\mu$ is the Lebesgue measure.
Throughout this work, for our experiments, we will be using the \textit{\hvilong{}} (\textit{\hvi{}}) metric as a cumulative performance indicator for \mo{} algorithms with respect to function evaluations. Given a new set of candidates $\gamma$, an existing Pareto set $\mathcal{P}$ and a reference point $r$, the \textit{\hvi{}} is formally defined as:
\begin{equation}
    HVI(P, r, \gamma) = \mathcal{H}_r(P \cup \gamma) - \mathcal{H}_r(P) \quad .
\end{equation}

\subsection{Multi-objective optimization for Deep Learning}
For \dl{}, it is often necessary to optimize not only the validation error (or validation accuracy) but also a cost metric, such as the inference time of a Neural Network or Floating Point Operations per Second. It is easy to imagine that a cost metric would be cheap to evaluate since it is a simple observation, unlike an objective such as accuracy (which would require the network to be trained first) \citep{izquierdo-icml21a}. Additionally, we might also be interested in a third objective like fairness or interpretability of the \dl{} model. However, from a \dl{} perspective, optimizing predictive performance typically (but not always) comes at the cost of degrading other objectives. Furthermore, benchmarks with more than three objectives are known to have highly correlating objectives \citep{li2021hwnasbench, bansal-jahs-neurips22a, dou2023eahasbench, NEURIPS2024_6ffdbf06}. In the context of Machine Learning, multi-objective algorithms for \hpo{} \citep{jin_multi-objective_2006} have been adapted mainly from the general \mo{} literature.

\paragraph{Scalarization-based \bolong{}}

Scalarization-based \molong{} \bolong{} (\mobo{}) approaches \citep{knowls-evoco06a, golovin_random_2020, paria_flexible_2019, yoon_sequential_2009} use a function: $s : \mathbb{R}^n \times \alpha \mapsto \mathbb{R} $ that maps the vector-valued \mo{} function into a scalar value, thus effectively converting the \mo{} problem into a \soolong{} problem. These approaches vary in the choice of the scalarization function \citep{knowls-evoco06a} or the distribution from which the weights are sampled \citep{yoon_sequential_2009, paria_flexible_2019}. \citet{knowls-evoco06a} introduced \textit{\parego{}}, which uses a Tchebycheff norm over the objective values as opposed to a linear weighted sum approach. These methods are highly scalable and easy to implement, which is why we employ random scalarizations during the \bo{} phase of \algo{}.

\paragraph{\mobolong{} using acquisition function modifications}

Other \mobo{} approaches directly modify the acquisition function in \bo{} to account for multiple objectives. \citet{emmerich-05a} proposed the \ehvilong{} (\ehvi{}) acquisition function wherein a surrogate model is fitted for each objective separately, and then the \eilong{} (\ei{}) \citep{jones-jgo98a} of the \hv{} contribution is calculated. Several improvements to calculate \ehvi{} have been proposed, such as in \citet{yang_multi-objective_2019} and \citet{daulton_differentiable_2020}. \ehvi{} is also used in \citet{ozaki-gecco20a} to extend the TPE \citep{bergstra-nips11a} to \mo{} TPE.
\citet{ponweiser_multiobjective_2008} introduced the S-Metric Selection-based EGO (SMS-EGO) which, instead of using \ehvi{}, selects new candidates by directly maximizing the \hv{} contribution based on the predictions of the surrogate model,  using the Lower Confidence Bound \citep{jones-jgo01a} acquisition function.
\citet{izquierdo-icml21a} modified \ehvi{} by fitting surrogate models only on the expensive objectives, such as validation accuracy.
\mo{} Information-theoretic acquisition functions, such as maximum entropy search \citep{belakaria-neurips19a} (MESMO) and predictive entropy search \citep{lobato-icml16a} (PESMO), aim to reduce the entropy of the location of the Pareto front.

\paragraph{Evolutionary algorithms}
Evolutionary \mo{} Algorithms mutate configurations from a diverse initial population to identify promising candidates closer to the Pareto front. \citet{deb_fast_2002} proposed the popular \nsgaiilong{} (\nsgaii{}) which uses \ndslong{} \citep{srinivas_muiltiobjective_1994} to rank candidates from multiple non-dominated fronts and conducts survival selection (tie-breaking) using \cdslong{} \citep{deb_fast_2002}.
\smsemoalong{} (\smsemoa{}) \citep{beume-or07a} also employs \ndslong{} from \citep{srinivas_muiltiobjective_1994} and \citep{deb_fast_2002} for the initial ranking of candidates, but then uses each candidate's contribution to the dominated \hv{} for survival selection. Evolutionary methods, however, are quite compute-inefficient, requiring a high budget to significantly improve the dominated \hv{}. Compute efficiency is one of the desiderata we identify in \tabref{table:desiderata} and therefore is a key aspect of \algo{}.

\paragraph{Multi-objective multi-fidelity optimization}

\citet{izquierdo-icml21a} extended \smsemoa{} to the \mf{} domain by augmenting it with \sh{} rungs. Furthermore, they introduced MO-\bohb{}, which replaced the TPE component with MO-TPE. \citet{schmucker-metalearn20a} adapted \hblong{} (\hb{}) to \mo{} using a randomly scalarized objective value (\hbrw{}) to select and promote promising configurations. 
\citet{salinas_multi-objective_2021} and \citet{schmucker-arxiv21a} further build on \citet{schmucker-metalearn20a} by modifying the promotion strategy of \hb{} and \asha{} respectively, using \ndslong{} for the initial ranking of candidates, and a greedy \epsnetlong{} (\epsnet{}) strategy for exploration.

MF-OSEMO \citep{belakaria_multi-fidelity_2020} and iMOCA \citep{belakaria2020informationtheoreticmultiobjectivebayesianoptimization} extend the information-theoretic method MESMO to discrete and continuous fidelities, respectively.
\citet{irshad_leveraging_2024} propose a novel modification to the \ehvi{} acquisition function which optimizes a multi-objective function and the fidelity of the data source jointly. They achieve this by defining a \textit{trust-based cost objective} which is directly proportional to the fidelity level. 
However, these \momf{}-\bo{} algorithms are quite computationally expensive, requiring vast amounts of resources and longer optimization runtimes. Although they integrate cheap approximations of the objective function, their high overall computational costs make them unsuitable for \dl{}.

Apart from a few notable exceptions, \mo{} algorithms have been largely been used for general optimization problems. Their usage in practical \dl{} applications have been relatively limited compared to \soolong{}, and only a handful of studies exist where \mo{} optimizers are benchmarked on real-world \dl{} tasks. We aim to bridge this gap between general \moolong{} and \mohpolong{} by demonstrating \algo{}'s effectiveness in both synthetic \mo{} problems, as well as \dl{} benchmarks.

\section{Algorithm details}\label{app:algo_details}


All the parts of \algo{} were implemented based on the \neps{} \citep{stoll_neural_2025} package. For \algo{}'s \init{} strategy we used \moasha{} already implemented in \neps{}. \moasha{} in \neps{} uses the \epsnet{} \mo{} promotion strategy from the \synetune{} repository, which is the original implementation by its authors \citep{schmucker-arxiv21a}. It is also important to note here that \moasha{} is the most viable choice for the \init{} compared to bandit-based optimizers with synchronous promotions like \hb{} or \sh{}. This is because the latter would require much longer budgets to promote configurations to the highest fidelity rung, which is impractical for the \init{} size of \bo{}. We set $\eta = 3$ and the \init{} size to 5.

We use the \botorch{} implementation of $\texttt{SingleTaskGP}$ as s surrogate model for purely continuous or purely discrete spaces, and $\texttt{MixedSingleTaskGP}$ for mixed search spaces. For the base \acqfunc{} in the prior-augmented \bo{}, we used $\texttt{qLogNoisy\ei{}}$ from \botorch{} as it has been proven to significantly outperform ordinary \ei{} implementations \citep{ament_unexpected_2023}. The \neps{} package already contains code for the $\texttt{WeightedAcquisition}$ function for \pibo{}, which we borrow for \algo{}.



\section{Sources of priors}\label{app_sec:prior_sources}
In practice, Deep Learning experts often have substantial intuition about ranges of hyperparameter settings that work best for certain models and downstream tasks \citep{souza-ecml21a, hvarfner-iclr22a, mallik-neurips23a}. For example, an ideal learning rate for the Adam optimizer \citep{kingma-iclr15a} is often around $1 \times 10^{-3}$ \citep{hvarfner-iclr22a}. Prior knowledge often stems from the literature, experience in tuning similar models, repeated experimentation or transfer learning across tasks. Consequently, \dl{} researchers often start their experiments with hyperparameter settings that are reported in the original paper for that architecture. In recent literature, large language models have also been used to generated priors for Hyperparameter Optimization \citep{zhang_llm_hpo_priors_2023_neurips_fmdm}. Priors for resource-related objectives, such as training time, FLOPS, memory consumption, or energy usage, are generally easier to define, as these objectives are directly observable and can hence be easily measured.

\section{Prior construction}\label{app_sec:study_priors}

For the construction of priors, we closely follow the procedure described by \citet{mallik-neurips23a}. Our priors are \hp{} settings, perturbed by a Gaussian noise with a $\sigma$ depending on the prior quality.  In all our experiments we use two kinds of priors for every objective - \textit{good} and \textit{bad} priors. The good priors represent areas of the \hp{} space where we expect the corresponding objective to have a value close to its \textit{optimum}.
The bad priors represent \textit{inaccurate} configurations which yield poor values for the objective function. The \hp{} configurations for these priors are generated using the methods listed below:
\begin{itemize}
    \item Class "\textbf{good}" priors: To generate \textit{good} priors, we begin by uniformly sampling 100,000 \hp{} configurations at random using a fixed global seed for all prior generation runs. We then evaluate these configurations on the corresponding benchmark at the highest available fidelity, $\maxfid{}$. Afterwards, we rank the configurations based on the objective values derived from their evaluations. Since we always aim to minimize each objective, for objectives intended to be maximized, we take their negative values to find the minimum. The configuration that yields the best objective value is perturbed by a Gaussian noise with $\sigma=0.01$.  This slight perturbation reflects a realistic scenario where prior knowledge is good or near-optimal, but never precisely so.
    \item Class "\textbf{bad}" priors: Similar to the good prior case, for the \textit{bad} priors, we sort the configurations based on the corresponding objective value. From this, we select the configuration with the worst seen value and do not perturb it any further. This forms our \textit{bad} prior configuration.
\end{itemize}
After locating the \hp{} configurations that constitute these priors, we create a Gaussian distribution over each, $\mathcal{N}(\lambda, \sigma^2)$, where $\sigma$ = 0.25 for all priors.

\section{Baselines}\label{app_sec:study_opts}



The implementation and hyperparameter setting of all baselines used in this paper are individually detailed below

\subsection{Single-objective baselines}

\paragraph{\bolong{} (\bo{})}\label{app_sec:base_bo}
\bolong{} is a popular \hpo{} algorithm that builds a probabilistic model to estimate the optimum of a blackbox objective function. We select the Gaussian Processes-based \bo{} implementation from the \neps{} \citep{stoll_neural_2025} package, which uses the $\texttt{q-Log-Noisy Expected Improvement}$ acquisition function from \botorch{}. This has been shown to perform significantly better than ordinary \eilong{} implementations. The \init{} size of the \bo{} is set to be the same as the dimensionality of the corresponding benchmark's search space \citep{ament_unexpected_2023}.

\paragraph{\hblong{} (\hb{})}
\hblong{} \citep{li-iclr17a} is a common multi-armed bandit-based \hpo{} algorithm that iterates over multiple \shlong{} brackets, and is a common baseline \mf{} benchmarking studies. The \neps{} package provides an implementation of \hb{} which allows for continuations, and we set $\eta=3$ for all our experiments.

\paragraph{\pibo{}}
\pibo{} is a \solong{}, blackbox optimization algorithm which augments the acquisition function with user-specified priors. We use the \pibo{} implementation from the \neps{} package. The original \pibo{} paper \citep{hvarfner-iclr22a} uses $\gamma = \frac{\beta}{n}$ to denote the power to which the prior PDF term is raised when multiplied by the values of the acquisition function, where $n$ refers to the \textit{n}-th iteration and the value of $\beta$ is set to 10. In the \neps{} package, however, $\gamma$ is completely different and is set to $e^{-n_{BO}/n_{d}}$, where $n_{BO}$ refers to the number of \bo{} samples and $n_{d}$ indicates the dimensions of the search space.

\subsection{Multi-objective baselines from the literature}

\paragraph{\borwlong{} (\borw{})}\label{app_sec:base_borw}

\bo{} with random weights is a popular \mo{} baseline which converts the \mo{} function into a \so{} optimization problem. Keeping all the settings as described in \bo{} above, we extend the \bo{} implementation in the \neps{} package by scalarizing the multivariate objective function $f$ with randomly chosen weights at every iteration of optimization.

\paragraph{\parego{}}

Just like \borw{}, \parego{} \citep{knowls-evoco06a} is another \bo{} baseline with the Chebyshev norm as the scalarization function. We use the \parego{} implementation from the \smac{} package and leave the \init{} design size of the \bo{} as the package default (search space dimensions).

\paragraph{\bolong{} with Expected Hypervolume Improvement acquisition function (\lnehvi{})}
The Expected Hypervolume Improvement acquisition function was introduced by \citep{emmerich-05a} which fits a surrogate model separately for each objective of the multi-objective problem. We employ the \botorch{} implementation of \lnehvilong{} \citep{ament_unexpected_2023} from the \neps{} package. The initial design size for \lnehvi{} is set to the size of the search space.

\paragraph{Multi-objective Tree-structured Parzen Estimator (\motpe{})}
\motpe{} \citep{ozaki-gecco20a, ozaki_multiobjective_2022} is an extension of the popular Tree-structure Parzen Estimator algorithm using \ehvi{} as an acquisition function. We use the Optuna implementation of \motpe{} with recent computational and performance improvements \citep{abe2025treestructuredparzenestimatorsolve}. Like the other \bo{} algorithms, we set the initial random sampling of Optuna's \motpe{} to the number of dimensions of the search space.

\paragraph{\nsgaii{}}

\nsgaii{} \citep{deb_fast_2002} is an EA algorithm which uses \ndslong{} to identify promising configurations and \cdslong{} as a tie-breaker. It is a popular baseline but EAs are quite sample-inefficient and hence not super practical for \dl{} as a standalone optimization algorithm. Thus, we use \nsgaii{} as a representative EA baseline and borrow its implementation from the \nvg{} package. The parameters of the algorithm are set to the defaults values defined in \nvg{}.

\paragraph{\momfbolong{}}

We incorporate a \momfbolong{} (\momfbo{}) from \citep{irshad_leveraging_2024} which uses a \textit{trust-based} cost objective which is directly proportional to the fidelity, optimizing the \molong{} function and the fidelity of the data source jointly. However, it is quite computationally expensive and running a single seed of this \momfbo{} over the \translate{} benchmark often took several hours.

\paragraph{\hbrwlong{} (\hbrw{})}

Following \citet{schmucker-metalearn20a}, we modify \hb{} from \neps{} with random weights the same way as \borw{} above. For all our experiments, we set the $\eta=3$.

\noindent
\begin{minipage}[tbp]{0.52\textwidth}
    \begin{algorithm}[H]
        \caption{Multi-objective selector}
        \label{algo:nondominated}
        \scriptsize
        \begin{algorithmic}[1]
            \Function{\textnormal{\texttt{non\_dom\_sorting}}}{$P$}
                \State $i = 1$
                \While{$P \neq \emptyset$}
                    \State $F_i \gets \{ \vec{\theta} \in P: \vec{\theta} \,\, \text{non-dominated} \}$
                    \State $(P, \, i) \gets (P \setminus F_i, \, i + 1)$
                \EndWhile
                \State \Return $F_1, \dots, F_m$
            \EndFunction

            \Statex \hrulefill

            \Function{\textnormal{\texttt{selector\_eps\_net}}}{$P$, $n$}
                \State $F_1, \dots, F_m \gets \texttt{non\_dom\_sorting}(P)$
                \State $\mathcal{C} \gets [ F_1\text{.pop}() ]$
                \For{$F = F_1, \dots, F_m$}
                    \While{$F \neq \emptyset$}
                        \State $\vec{\theta} \gets \min_{\vec{\theta} \in F} \max_{\vec{\theta}' \in \mathcal{C}} \| \vec{y}_{\vec{\theta}} - \vec{y}_{\vec{\theta}'} \|_2$
                        \State $\mathcal{C}\text{.append}(\vec{\theta})$
                        \State $F \gets F \setminus \{\vec{\theta}\}$
                    \EndWhile
                \EndFor
                \State \Return $\mathcal{C}_1, \dots, \mathcal{C}_n$
            \EndFunction

            \Statex \hrulefill

            \Function{\textnormal{\texttt{selector\_nsga\_ii}}}{$P$, $n$}
                \State $F_1, \dots, F_m \gets \texttt{non\_dom\_sorting}(P)$
                \State $\mathcal{C} \gets [ \, ]$
                \For{$F = F_1, \dots, F_m$}
                    \State $\mathcal{C} \gets \mathcal{C} + \texttt{crowding\_dist\_sorting}(F)$
                \EndFor
                \State \Return $\mathcal{C}_1, \dots, \mathcal{C}_n$
            \EndFunction
        \end{algorithmic}
    \end{algorithm}
\end{minipage}%
\hfill
\begin{minipage}[tbp]{0.46\textwidth}
    \begin{algorithm}[H]
        \caption{Multi-objective ASHA}
        \label{algo:mo-asha}
        \scriptsize
        \begin{algorithmic}[1]
            \State \textbf{Input:} $R$, $r_0$, $s$, $\eta$ (default $\eta = 3$)

            \Function{\textnormal{\texttt{mo\_asha}}}{}
                \Repeat
                    \For{each free worker}
                        \State $(\vec{\theta}, k) \gets \texttt{get\_job}()$
                        \State $\texttt{run\_evaluation}(\vec{\theta}, r_0 \eta^{s + k})$
                    \EndFor
                    \For{completed job $(\vec{\theta}, k)$ with objective vector $\vec{y}_{\vec{\theta}}$}
                        \State update $\vec{\theta}$ in rung $k$ with $\vec{y}_{\vec{\theta}}$
                    \EndFor
                \Until{budget exhausted}
            \EndFunction

            \Statex \hrulefill

            \Function{\textnormal{\texttt{get\_job}}}{}
                \For{$k = \lfloor \log_\eta(R / r_0) \rfloor - s - 1, \dots, 1, 0$}
                    \State $\text{candidates} \gets $ \\ $\texttt{mo\_selector}(\text{rung} \, k, \left| \text{rung} \, k \right| / \eta)$
                    \State $\text{promotable} \gets $ \\ $\{ \vec{\theta} \in \text{candidates}: \vec{\theta} \,\, \text{not promoted} \}$
                    \If{$\, | \text{promotable} | > 0$}
                        \State \Return (promotable[0], $k + 1$)
                    \EndIf
                \EndFor
                \State draw random configuration $\vec{\theta}$
                \State \Return $(\vec{\theta}, 0)$
            \EndFunction
        \end{algorithmic}
    \end{algorithm}
\end{minipage}

\paragraph{\moashalong{} (\moasha{})}

\moasha{} \citep{schmucker-arxiv21a} (Algorithm~\ref{algo:nondominated}, \ref{algo:mo-asha}) is an infinite horizon \mo{} optimizer and currently one of the \sota{} baselines for \moolong{}, using bandit-based \asha{} as the base. Like \asha{}, \moasha{} can also run very efficiently on \hpo{} setups with many parallel workers, reducing idle-time. However, even for single worker setups, \moasha{} is able to leverage its asynchronous promotion strategy to achieve competitive performance, and that is what we employ for the experiments in this paper. We used \moasha{} from the \neps{} package, and just like \hbrw{} above, we set $\eta = 3$.

\subsection{Baselines we constructed to leverage multi-objective expert priors}

We modify single-objective approaches from the literature (\RSprior {}, \priormoasha{}, \pibo{} \citep{hvarfner-iclr22a}, \pb{} \citep{mallik-neurips23a}) to randomly chose and sample from one of the multi-objective priors at each iteration. We further augment \pibo{} with random scalarizations and modify \pb{}'s ensemble sampling policy using scalarized incumbents for \mo{} to build \mopb{}.

\paragraph{\piborwlong{} (\piborw{})}

 \piborw{} is a \mo{} direct extension of \pibo{} from \neps{} with random weights, just like \borw{} above. For use with \mo{} priors, we modify \pibo{} to randomly chose and sample from one of the \mo{} priors at each iteration.  Like \borw{}, we set the \init{} size to $n_{d}$, and sample from a randomly chosen prior for each of the initial points. The remaining details of the base \pibo{} algorithm is the same as detailed above.

\paragraph{\pb{}+\bo{}}
\pb{} integrates cheap proxies unlike \pibo{} to achieve good anytime performance. It employs an ESP strategy for sampling proportionately from the \textit{priors}, the \textit{incumbent} and at \textit{random}. \pb{}+\bo{} is a model-based extension of \pb{} using Gaussian Processes with the \ei{} acquisition function. \neps{} provides an implementation of \pb{}+\bo{} which we use for our single-objective experiments. Just like in \bo{} above, \pb{}+\bo{} uses the $\texttt{q-Log-Noisy Expected Improvement}$ acquisition function from \botorch{}. Further details about this model-based extension is available in the \pb{} paper \citep{mallik-neurips23a}.

\paragraph{\mopb{}}

We extend \pb{} to the \mo{} domain by first replacing the \mf{} component with an \momf{} component. Then, to calculate the \topk{} configurations, we scalarize the \mo{} vectors using weights, randomly chosen during each iteration. Additionally, to integrate \mo{} priors, \mopb{} chooses one of the available priors at random at each iteration. We note that a scalarization-based incumbent modification works better for \mopb{} than a Pareto front incumbent such as \epsnet{}. Additionally, we set \mopb{}'s $\eta = 3$, just as in \pb{}.
 
\paragraph{\RSpriorlong{} (\RSprior{})}

We first randomly select one of the \molong{} priors to sample from. We then equip \RSlong{} to draw random samples from the selected prior distribution (instead of the entire search space) to construct the \RSprior{} baseline.

\paragraph{\moasha{} + Prior}
We equip \moasha{} to sample from one of the \molong{} prior distributions in the same way as \RSlong{} above. For \moasha{} + 50\% prior sampling, we choose uniformly at random, between prior-based sampling and random sampling from the entire search space.

\section{Benchmarks}\label{app_sec:study_benches}

Our main experiments in \Secref{sec:exp} include the surrogate benchmarks \lcba{}, \lcbb{}, \lcbc{}, \lcbd{} from \yahpogym{} and \cifar{}, \imagenet{}, \translate{}, \lmbt{} from the \pdone{} suite. For an overview of the search spaces of all our benchmarks, see \tabref{tab:benches_overview}.

\begin{table}[t]
\centering
\caption{Overview of benchmark search spaces.}
\label{tab:benches_overview}
\begin{tabular}{lccc}
\toprule
Benchmark & \# Total HPs & \# Continuous HPs & \# Discrete HPs \\
\midrule
\texttt{cifar-100-wide\_resnet-2048} & 5 & 4 & 1 \\
\texttt{imagenet-resnet-512} & 5 & 4 & 1 \\
\texttt{lm1b-transformer-2048} & 5 & 4 & 1 \\
\texttt{translatewmt-xformer-64} & 5 & 4 & 1 \\
\texttt{yahpo-lcbench 126026} & 7 & 4 & 3 \\
\texttt{yahpo-lcbench 146212} & 7 & 4 & 3 \\
\texttt{yahpo-lcbench 168330} & 7 & 4 & 3 \\
\texttt{yahpo-lcbench 168868} & 7 & 4 & 3 \\
\bottomrule
\end{tabular}
\end{table}

\subsection{PD1 (HyperBO)}

PD1 from HyperBO \citep{wang_pre-trained_2024} is a collection of XGBoost surrogates trained on the learning curves of near \sota{} \dl{} models on a diverse array of practical downstream \dl{} tasks including image classification, language modeling and language translation. 
Overall, PD1 contains 24 benchmarking tasks, with each consisting of a task dataset, a \dl{} model, and a broad search space for Nesterov Momentum \citep{nesterov-smd83a}.

From these 24, we select 4 benchmarks from $\texttt{mf-prior-bench}$ \citep{Bergman_mf-prior-bench_2025} providing a well-rounded representation of \dl{} models and the aforementioned tasks. For each of these benchmarks, we select the $\texttt{valid\_error\_rate}$ as the \textit{validation error} objective and $\texttt{train\_cost}$ as the \textit{training cost} objective. All of these benchmarks have a single fidelity $\texttt{epoch}$. We list the static reference points for calculating the \hvi{} for the \pdone{} benchmarks in \tabref{table:pd1_max_bounds}. The individual benchmarks are further detailed below:
\begin{enumerate}
    \item \textbf{cifar100-wide\_resnet-2048} benchmark contains the optimization trace of a $\texttt{WideResnet}$ \citep{zagoruyko-bmvc16a} model on the $\texttt{CIFAR-100}$ \citep{krizhevsky-tech09a} dataset with a batch size of 2048. The \hp{} space of this benchmark is given in \tabref{table:cifar100_hps}.
    \item \textbf{imagenet-resnet-512} surrogate is trained on the learning curve of a $\texttt{ResNet50}$ \citep{he-cvpr16a} on the $\texttt{ImageNet}$ \citep{russakovsky-ijcv15a} dataset with a batch size of 512. See \tabref{table:imagenet_hps} for the detailed search space of this benchmark.
    \item \textbf{lm1b-transformer-2048} is a surrogate trained on the \hpo{} runs of a transformer model \citep{roy-etal-2021-efficient} on the \textit{One Billion Word} statistical language modeling benchmark \citep{chelba2014billionwordbenchmarkmeasuring}. \tabref{table:lm1b_hps} lists the search space of the benchmark.
    \item \textbf{translatewmt-xformer-64} surrogate is trained on the \hpo{} runs of an $\texttt{xformer}$ \citep{xFormers2022} transformer model on the $\texttt{WMT15 German-English}$ text translation dataset \citep{bojar-etal-2015-findings}. For the detailed search space, see \tabref{table:translatewmt_hps}.
\end{enumerate}

\begin{table}[!ht]
  \caption[Reference points for PD1 benchmarks.]{Reference values for $\texttt{valid\_error\_rate}$ and $\texttt{train\_cost}$ objectives across PD1 benchmarks for \hvi{} calculation.}
  \label{table:pd1_max_bounds}
  \centering
  \small
  \begin{tabular}{p{5cm}cc}
    \toprule
    Benchmark Name & $\texttt{valid\_error\_rate}$ (max) & $\texttt{train\_cost}$ (max) \\
    \midrule
    $\texttt{cifar100\text{-}wide\_resnet\text{-}2048}$     & 1.0 & 30 \\
    $\texttt{imagenet\text{-}resnet\text{-}512}$           & 1.0 & 5000 \\
    $\texttt{lm1b\text{-}transformer\text{-}2048}$         & 1.0 & 1000 \\
    $\texttt{translatewmt\text{-}xformer\text{-}64}$       & 1.0 & 20000 \\
    \bottomrule
  \end{tabular}
\end{table}

\begin{table}[!ht]
  \caption[Hyperparameter search space for PD1 CIFAR-100 benchmark.]{Hyperparameter search space table of the $\texttt{cifar-100-wide\_resnet-2048}$ benchmark, including the \hp{} ranges and fidelity bounds of $\texttt{epoch}$, as given in $\mfpbench{}$.}
  \label{table:cifar100_hps}
  \centering
  \small
  \begin{tabular}{p{3cm}lcccc}
    \toprule
    Hyperparameter & Type & Log-scaled & Range & Space Type & Notes \\
    \midrule
    $\texttt{lr\_decay\_factor}$ & float &  & $[0.010093, 0.989012]$ & continuous & \\
    $\texttt{lr\_initial}$ & float & \cmark & $[0.000010, 9.779176]$ & continuous & \\
    $\texttt{lr\_power}$ & float &  & $[0.100708, 1.999376]$ & continuous & \\
    $\texttt{opt\_momentum}$ & float & \cmark & $[0.000059, 0.998993]$ & continuous & \\
    \midrule
    $\texttt{epoch}$   & integer &  & $[1, 52]$   & discrete   & fidelity \\
    \bottomrule
  \end{tabular}
\end{table}

\begin{table}[!ht]
  \caption[Hyperparameter search space for PD1 ImageNet benchmark.]{Approximate hyperparameter search space table of the $\texttt{imagenet-resnet-512}$ benchmark, including \hp{} ranges and fidelity bounds of $\texttt{epoch}$. Exact ranges are provided by $\mfpbench{}$.}
  \label{table:imagenet_hps}
  \centering
  \small
  \begin{tabular}{p{3cm}lcccc}
    \toprule
    Hyperparameter & Type & Log-scaled & Range & Space Type & Notes \\
    \midrule
    $\texttt{lr\_decay\_factor}$ & float &   & $[0.010294, 0.989753]$ & continuous &  \\
    $\texttt{lr\_initial}$ & float & \cmark & $[1e{-}5, 9.774312]$ & continuous &  \\
    $\texttt{lr\_power}$ & float &   & $[0.100225, 1.999326]$ & continuous &  \\
    $\texttt{opt\_momentum}$ & float & \cmark & $[5.9e{-}5, 0.998993]$ & continuous &  \\
    \midrule
    $\texttt{epoch}$   & integer &   & $[1, 99]$   & discrete   & fidelity  \\
    \bottomrule
  \end{tabular}
\end{table}

\begin{table}[!ht]
  \caption[Hyperparameter search space for PD1 LM1B-Transformer benchmark.]{Hyperparameter search space of the $\texttt{lm1b-transformer-2048}$ benchmark, with the fidelity $\texttt{epoch}$ as given in $\mfpbench{}$.}
  \label{table:lm1b_hps}
  \centering
  \small
  \begin{tabular}{p{3cm}lcccc}
    \toprule
    Hyperparameter & Type & Log-scaled & Range & Space Type & Notes \\
    \midrule
    $\texttt{lr\_decay\_factor}$ & float &   & $[0.010543, 0.9885653]$ & continuous & \\
    $\texttt{lr\_initial}$ & float & \cmark & $[1e{-}5, 9.986256]$ & continuous & \\
    $\texttt{lr\_power}$ & float &   & $[0.100811, 1.999659]$ & continuous & \\
    $\texttt{opt\_momentum}$ & float & \cmark & $[5.9e{-}5, 0.9989986]$ & continuous & \\
    \midrule
    $\texttt{epoch}$   & integer &   & $[1, 74]$   & discrete   & fidelity \\
    \bottomrule
  \end{tabular}
\end{table}

\begin{table}[!ht]
  \caption[Hyperparameter search space for PD1 Translate-WMT-xformer benchmark.]{Search space and fidelity $\texttt{epoch}$ of the $\texttt{translatewmt-xformer-64}$ benchmark, as given in $\mfpbench{}$.}
  \label{table:translatewmt_hps}
  \centering
  \small
  \begin{tabular}{p{3cm}lcccc}
    \toprule
    Hyperparameter & Type & Log-scaled & Range & Space Type & Notes \\
    \midrule
    $\texttt{lr\_decay\_factor}$ & float &   & $[0.010022, 0.988565]$ & continuous & \\
    $\texttt{lr\_initial}$ & float & \cmark & $[1.00276e{-}5, 9.842248]$ & continuous & \\
    $\texttt{lr\_power}$ & float &   & $[0.100425, 1.998593]$ & continuous & \\
    $\texttt{opt\_momentum}$ & float & \cmark & $[5.86114e{-}5, 0.998999]$ & continuous & \\
    \midrule
    $\texttt{epoch}$   & integer &   & $[1, 19]$   & discrete   & fidelity \\
    \bottomrule
  \end{tabular}
\end{table}

\subsection{LCBench surrogate benchmarks (YAHPO-Gym)}

\yahpogym{} \citep{pfisterer_yahpo_2022} is a large collection of \molong{} \mflong{} surrogate benchmarks trained on a wide array of tasks with fidelities including epochs as well as dataset fractions. \yahpogym{} also contains surrogates for the \lcbench{} \citep{zimmer-tpami21a} set of benchmarks that consists of surrogates trained on the learning curves of \dl{} models, on several OpenML \citep{vanschoren-sigkdd14a} datasets. Out of these, we choose 4 task OpenML IDs for the experiments in this paper - 126026, 146212, 168330 and 168868. The fidelity for these tasks is $\texttt{epoch}$ and we select the $\texttt{val\_cross\_entropy}$ and $\texttt{time}$ as the \textit{validation error} and the \textit{training cost} objectives respectively, for our experiments. \tabref{table:lcbench_max_bounds} lists the maximum bounds used as the reference points for calculating the \hvilong{}, for each of the selected \lcbench{} task IDs. All \lcbench{} benchmarks share a common search space, detailed in \tabref{table:lcbench_hps}.

\begin{table}[!ht]
  \caption[Reference points for the YAHPO-Gym LCBench benchmarks.]{Reference values for $\texttt{val\_cross\_entropy}$ and $\texttt{time}$ objectives across selected LCBench tasks, for \hvi{} calculation.}
  \label{table:lcbench_max_bounds}
  \centering
  \begin{tabular}{rcc}
    \toprule
    Task ID & $\texttt{val\_cross\_entropy}$ (max) & $\texttt{time}$ (max, seconds) \\
    \midrule
    126026 & 1.0 & 150 \\
    146212 & 1.0 & 150 \\
    168330 & 1.0 & 5000 \\
    168868 & 1.0 & 200\\
    \bottomrule
  \end{tabular}
\end{table}

\begin{table}[!ht]
  \caption[Hyperparameter search space for YAHPO-Gym LCBench benchmarks.]{Hyperparameter search space table of the $\texttt{yahpo-lcbench}$ benchmarks. This includes the \hp{} ranges and types as typically defined in the YAHPO-Gym benchmark suite.}
  \label{table:lcbench_hps}
  \centering
  \small
  \begin{tabular}{p{3cm}lcccc}
    \toprule
    Hyperparameter & Type & Log-scaled & Range & Space Type & Notes \\
    \midrule
    $\texttt{batch\_size}$   & integer & \cmark & $[16, 512]$     & discrete  & \\
    $\texttt{learning\_rate}$& float   & \cmark & $[1\mathrm{e}{-4}, 0.1]$ & continuous & \\
    $\texttt{momentum}$      & float   &        & $[0.1, 0.99]$   & continuous & \\
    $\texttt{weight\_decay}$ & float   &        & $[1\mathrm{e}{-5}, 0.1]$ & continuous & \\
    $\texttt{num\_layers}$   & integer &        & $[1, 5]$        & discrete  & \\
    $\texttt{max\_units}$    & integer & \cmark & $[64, 1024]$    & discrete  & \\
    $\texttt{max\_dropout}$  & float   &        & $[0.0, 1.0]$    & continuous & \\
    \midrule
    $\texttt{epoch}$  & integer   &        & $[1, 52]$    & discrete & fidelity \\
    \bottomrule
  \end{tabular}
\end{table}




\section{Details on the large-scale case study}\label{app_sec:real_world}

The OpenCLIP pipeline trains CLIP \citep{radford-arxiv21a} models using an open-source implementation \citep{Cherti_2023} on a 1M-sample subset of the LAION-400M dataset \citep{schuhmann2021laion400mopendatasetclipfiltered}. We determined that training on a 1M-sample subset is sufficient to observe meaningful performance differences on downstream tasks while remaining computationally tractable for optimization benchmarking. For our fidelity-based evaluation, we use training epochs (from 1 to 10) as our cheap proxy.

The hyperparameter search space comprises 5 parameters, including the fidelity dimension, which are detailed in \tabref{table:openclip_space}. During the optimization process, several architectural and optimizer parameters are kept constant to isolate the effects of the chosen search space: $\texttt{text\_heads} = 8$, weight decay ($\texttt{wd}$) $= 1.0\times 10^{-6}$, Adam optimizer betas ($\texttt{beta1} = 0.9$, $\texttt{beta2} = 0.999$), and $\texttt{eps} = 1.0\times 10^{-6}$.

We evaluate \algo{} against the strongest baselines, comprising \moasha{}, \borw{}, \motpe{}, and \RS{}, under a budget of 10 PFLOPs across 20 seeds. In \Figref{fig:vlm}, we plot the mean performance, with shaded regions representing the standard error.
\begin{table}[!ht]
  \caption{Hyperparameter search space for the OpenCLIP benchmark. The $\texttt{epochs}$ parameter serves as the fidelity for the optimization process.}
  \label{table:openclip_space}
  \centering
  \small
  \begin{tabular}{p{3.5cm}lcccc}
    \toprule
    Hyperparameter & Type & Log-scaled & Range/Set & Space Type & Notes \\
    \midrule
    \texttt{lr} & float & \checkmark & $[0.0001, 0.01]$ & continuous & \\
    \texttt{warmup\_fraction} & float & & $[0.0, 0.75]$ & continuous & \\
    \texttt{vision\_width} & integer & & $\{32, 64, 128, 192, 256\}$ & discrete & \\
    \texttt{text\_width} & integer & & $\{32, 64, 128, 192, 256\}$ & discrete &  \\
    \midrule
    \texttt{epochs} & integer & & $[1, 10]$ & discrete & fidelity \\
    \bottomrule
  \end{tabular}
\end{table}

\section{Details on evaluation protocol}\label{app_sec:study_setup}

\paragraph{Computing hypervolume} We compute the \hv{} with respect to a static reference point set for each benchmark (\appref{app_sec:study_benches}).

\paragraph{Equivalent function evaluations} For blackbox optimizers like \borw{}, \nsgaii{} and \parego{}, every optimization iteration is equal to a function evaluation since they evaluate $f$ at the maximum fidelity $\maxfid{}$. For optimizers such as \moasha{}, \hbrw{} and \algo{} that use cheap proxies of the objective, we calculate equivalent function evaluations as $z/\maxfid{}$ where $z$ is the fidelity at which $f$ is evaluated at a given iteration. We note here that for all \mf{} optimizers, we leveraged continuations and plot the \hv{} only when an equivalent full function evaluation has been performed, \ie{}, when the benchmark is evaluated at its highest fidelity $\maxfid{}$.

We note that for fairness, we generated the plots for \momfbo{} using a fidelity-aware trust threshold of 70\% of the respective benchmark's $\maxfid{}$, \ie{}, only evaluations at sufficiently high fidelity were counted as informative progress, while lower-fidelity probes were treated as exploratory. This avoids penalizing \momfbo{} for its sampling strategy, which queries many fidelities before concentrating on the most trusted evaluations.

\paragraph{Single-objective evaluations} We report the relative rankings of all optimizers over all benchmarks based on the normalized regret per benchmark. Similar to the multi-objective case, we run each optimizer-benchmark pair for 20 equivalent full function evaluations, across 25 random seeds.


\section{Ablation study of the BO components of \algo{}}\label{app:bo_ablation}

In this section, we perform an ablation study of the scalarization method and acquisition function of \algo{}'s \eps{}-\bo{}. In particular, we create 3 different versions of \algo{} using Hypervolume scalarization (HV) \citep{golovin_random_2020}, and the Lower Confidence Bound (LCB) \citep{auer_using_2002, auer_finite-time_2002} and Thompson Sampling (TS) \citep{thompson-biomet33a} acquisition functions, to justify our design choice and answer \rqf{}. As noted in \appref{app:algo_details}, \algo{} uses random linear scalarization and the Expected Improvement acquisition function. We note here that we only ablate one component at a time, \ie{}, ablations using a different acquisition function still use linear scalarization, while the ablation using the Hypervolume scalarization uses the Expected Improvement acquisition function. 

\begin{figure}[t]
    \centering
    \includegraphics[width=1.0\linewidth]{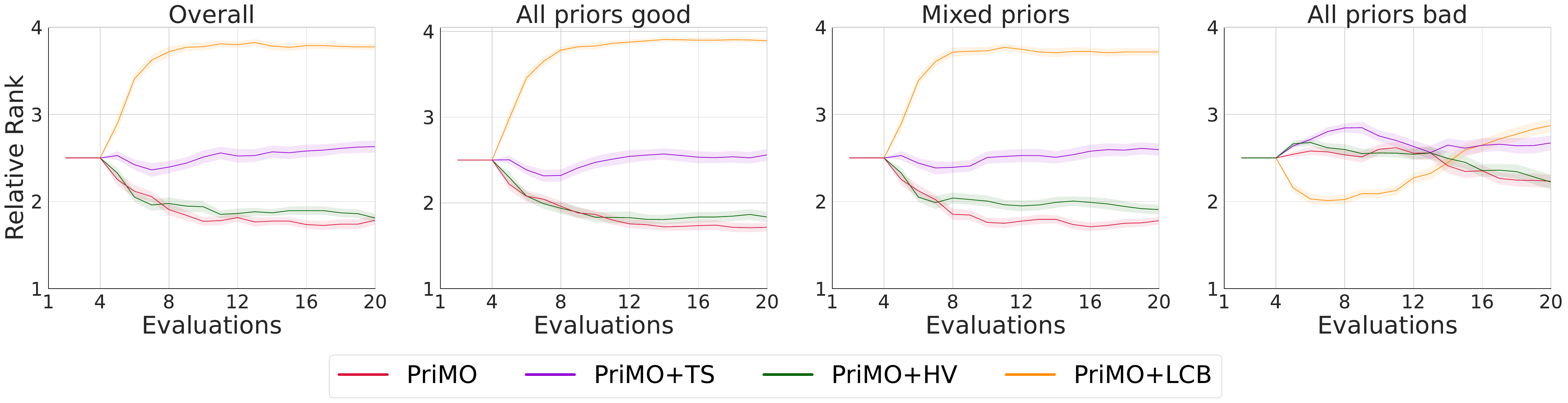}
    \caption{Mean relative ranks $\pm$ 1 standard error of various ablations of the \eps{}-\bo{} of \algo{} across benchmarks and seeds under various prior conditions.}
    \label{fig:ovranks_ablations_bo_all}
\end{figure}

\Figref{fig:ovranks_ablations_bo_all} shows that after the end of the \init{} strategy, the acquisition function has a significant impact on the relative ranking of the ablations, while the effect of the scalarization method is marginal. Overall, the \algo{} ablation with the Lower Confidence Bound acquisition function is the worst across all prior conditions, followed by Thompson Sampling. Hypervolume scalarization performs well and is marginally worse than the original design of \algo{} under all prior conditions. However, \algo{} remains the best design overall and under all prior conditions, thereby justifying our choices for the BO components of \algo{} (\rqf{}).

\section{\eps{} sensitivity study}\label{app:eps_study}

\begin{figure}[t]
    \centering
    \includegraphics[width=1.0\linewidth]{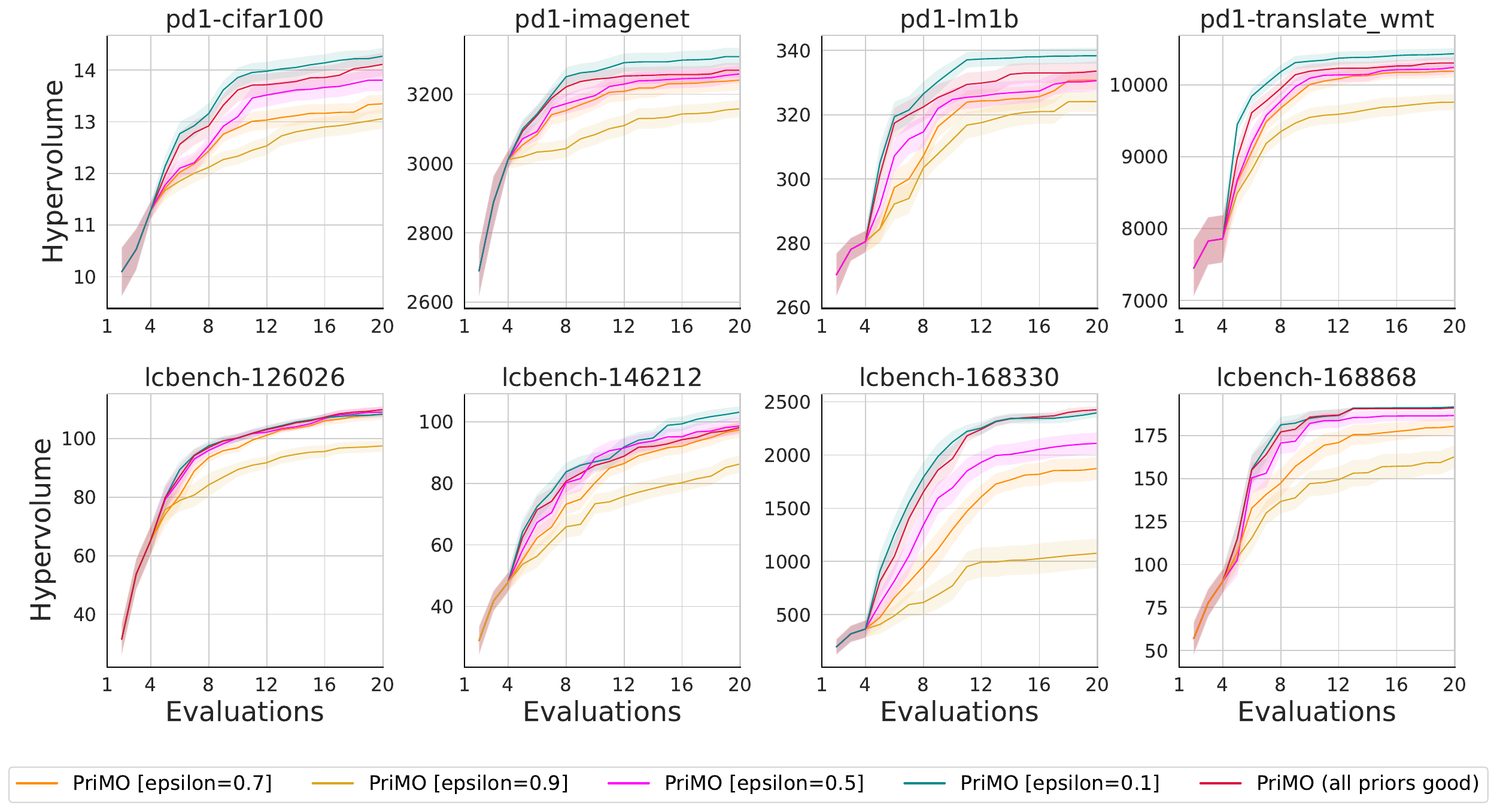}
    \caption{Sensitivity study of \eps{} in \algo{}'s \eps{}-\bo{} using mean dominated \hvlong{} $\pm$ 1 standard error across seeds for each benchmark. \algo{} is under all good priors setting here.}
    \label{fig:hv_eps_study_all_good}
\end{figure}

In \Figref{fig:hv_eps_study_all_good}, we perform a sensitivity study on the \eps{} in \algo{}'s \bo{}, across multiple seeds and all benchmarks under good prior conditions. Note that \eps{} denotes the probability with which \algo{} select vanilla \bo{} \ie{} without the prior-augmentation in the acquisition function. A high value of \eps{} would indicate convergence towards vanilla \bolong{}, while an extremely low value would mean convergence towards \piborw{}. An \eps{}=0.5 would essentially mean a 'coin-toss' between prior-based and standard \bolong{}. We choose \eps{}=0.25 (in red) which appears to strike the perfect balance between exploiting prior knowledge and exploration without a prior, as well as, aiding in robustness in case of a misleading prior.

\section{Additional experiments and analysis}\label{app:all_exp}

In this section,
we present detailed \hvlong{} and Pareto plots across all 8 benchmarks under good and bad priors, comparing \algo{} against both, non-prior and prior-based MO-adapted baselines.

\subsection{Performance under good priors}\label{subsec:res_strong}

\begin{figure}[t]
    \centering
    \includegraphics[width=1.0\linewidth]{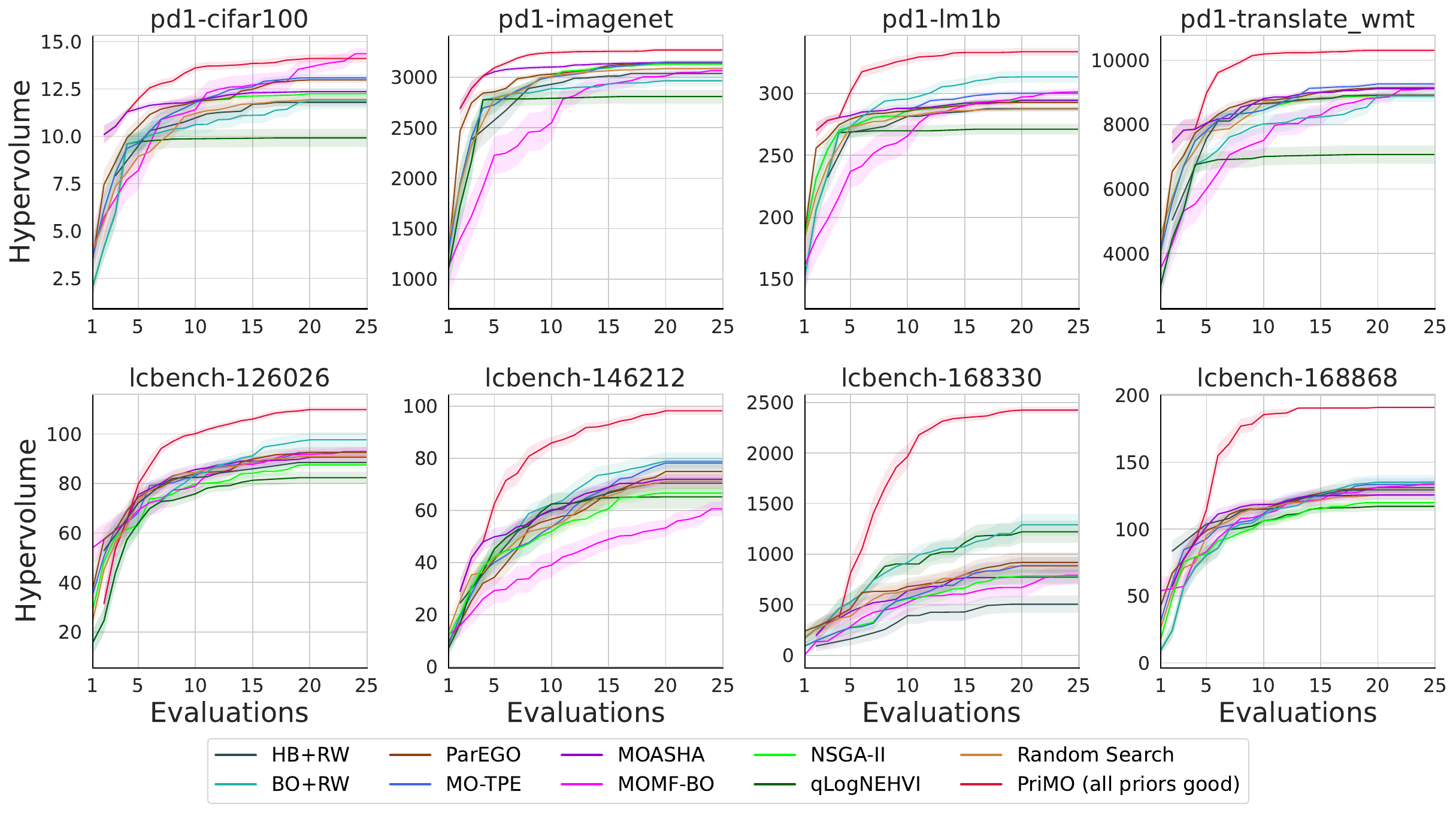}
    \caption{Mean dominated \hvlong{} $\pm$ 1 standard error of \algo{} and prominent multi-objective algorithms across seeds for each benchmark. \algo{} is under all good priors setting here.}
    \label{fig:hv_subset1_all_good}
\end{figure}

\paragraph{Hypervolume Plots across all benchmarks under good prior conditions against non-prior baselines}
\Figref{fig:hv_subset1_all_good} shows that \algo{} exhibits strong starts across all benchmarks, with respect to the mean dominated \hvlong{}, and is the best performing \hpo{} algorithm early on. As explained in \Secref{exp:sota}, we attribute this head start to the novel \init{} strategy which samples good seed points. The \eps{}-\bo further leverages these strong initial points to achieve \sota{} performance across all benchmarks, with the exception of \cifar{}, where \momfbo{} marginally outperforms \algo{} at the end of the budget.

\begin{figure}[tb]
    \centering
    \includegraphics[width=1\linewidth]{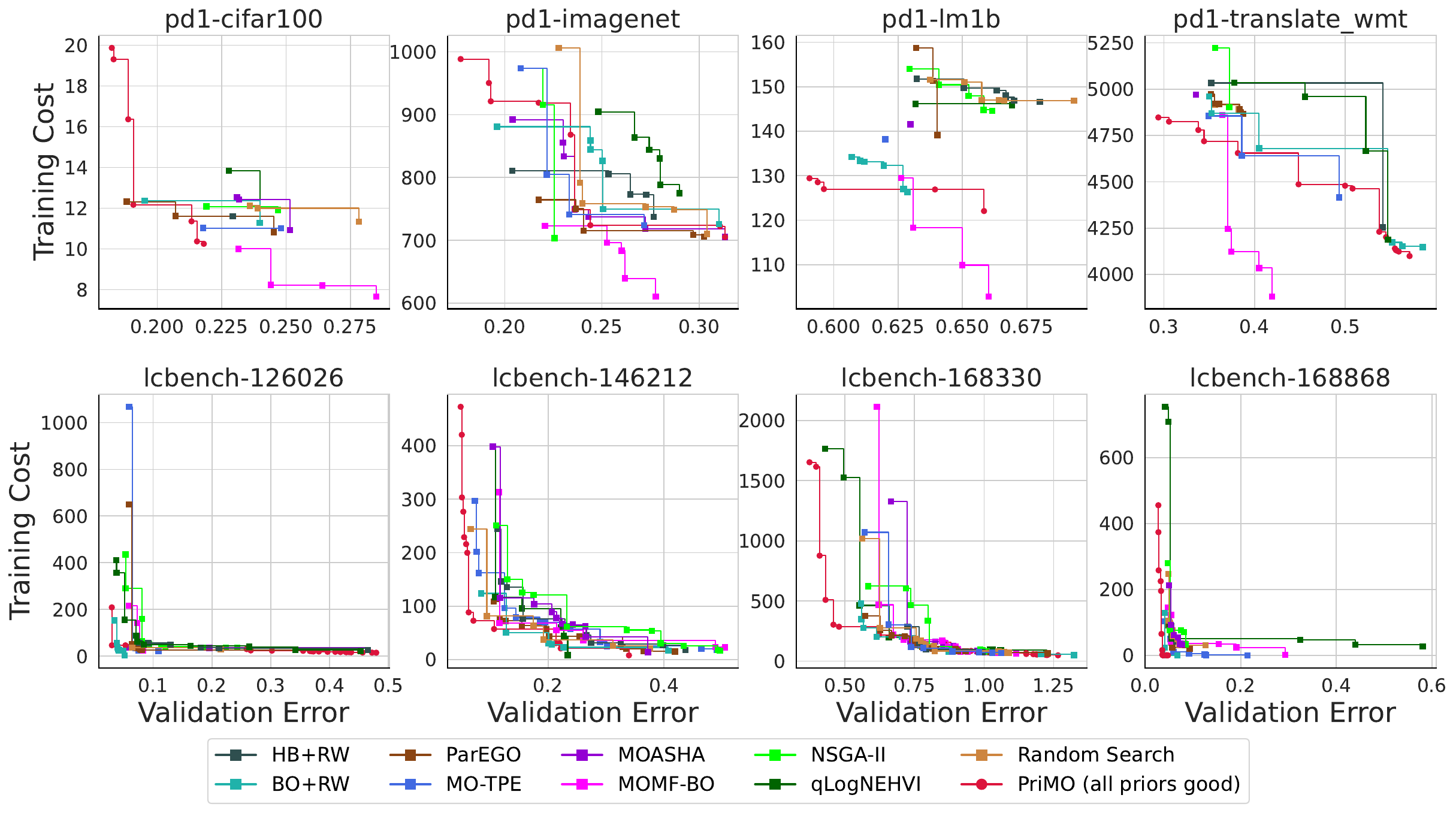}
    \caption{Shown here are the Pareto fronts obtained by \algo{}, compared to other non-prior \mo{} baselines under all good prior conditions.}
    \label{fig:pareto_subset1_all_good}
\end{figure}

\paragraph{Pareto Front analysis and comparison against non-prior baselines.}
For every baseline, we report the Pareto front aggregated across all seeds per benchmark in line with existing literature \citep{izquierdo-icml21a, schmucker-metalearn20a, schmucker-arxiv21a}, with the primary (validation error) objective on the x-axis and the training cost objective along the y-axis. The Pareto front plot in \Figref{fig:pareto_subset1_all_good} shows that, on average, \algo{} locates the most non-dominated points compared to the baselines on the benchmarks.

\paragraph{Case-level comparison across all benchmarks}
In \tabref{table:pareto_case_comp}, we perform a case-level analysis across all benchmarks and compare the performance of \algo{} and \borw{} based on the Pareto fronts aggregated across all seeds. We notice that, compared to standard \borw{}, \algo{} achieves lower validation error across all benchmarks and lower training cost in most. We further notice that on 5 out of 8 benchmarks, \algo{} clearly has \textbf{both} the lower error and model cost, further underlining \algo{}'s better coverage of the Pareto front.

\begin{table}[!ht]
\centering
\small
\caption{Case level comparison of performance of \algo{} against \borw{} on 8 \dllong{} benchmarks based on aggregated Pareto fronts.}
\label{table:pareto_case_comp}
\begin{tabular}{p{3.2cm}cccccc}
\toprule
Benchmark & 
\makecell{\algo{}\\\text{Best Error}} & 
\makecell{\borw{}\\\text{Best Error}} & 
\makecell{\algo{}\\\text{Min Cost}} & 
\makecell{\borw{}\\\text{Min Cost}} & 
\makecell{\text{Superior}\\\text{Algorithm}} \\
\midrule
\cifar{} & \textbf{0.185} & 0.188 & \textbf{10.46} & 11.27 & \textbf{\algo{}} \\
\imagenet{} & \textbf{0.178} & 0.218 & \textbf{696.0} & 707.0 & \textbf{\algo{}} \\
\lmbt{} & \textbf{0.586} & 0.605 & 119.5 & \textbf{117.9} & - \\
\translate{} & \textbf{0.303} & 0.347 & \textbf{4120} & 4140 & \textbf{\algo{}} \\
\lcba{} & \textbf{0.0307} & 0.0390 & 14.97 & \textbf{1.43} & -\\
\lcbb{} & \textbf{0.0509} & 0.0841 & \textbf{7.94} & 9.49 & \textbf{\algo{}} \\
\lcbc{} & \textbf{0.379} & 0.473 & 50.77 & \textbf{48.96} & - \\
\lcbd{} & \textbf{0.0284} & 0.0353 & \textbf{0.00154} & \textbf{0.00154} & \textbf{\algo{}} \\
\bottomrule
\end{tabular}
\end{table}

\paragraph{Performance against constructed prior-based \mo{} baselines under good priors}\label{app:prior_baselines}

\begin{figure*}[t]
    \centering
    \includegraphics[width=1.0\linewidth]{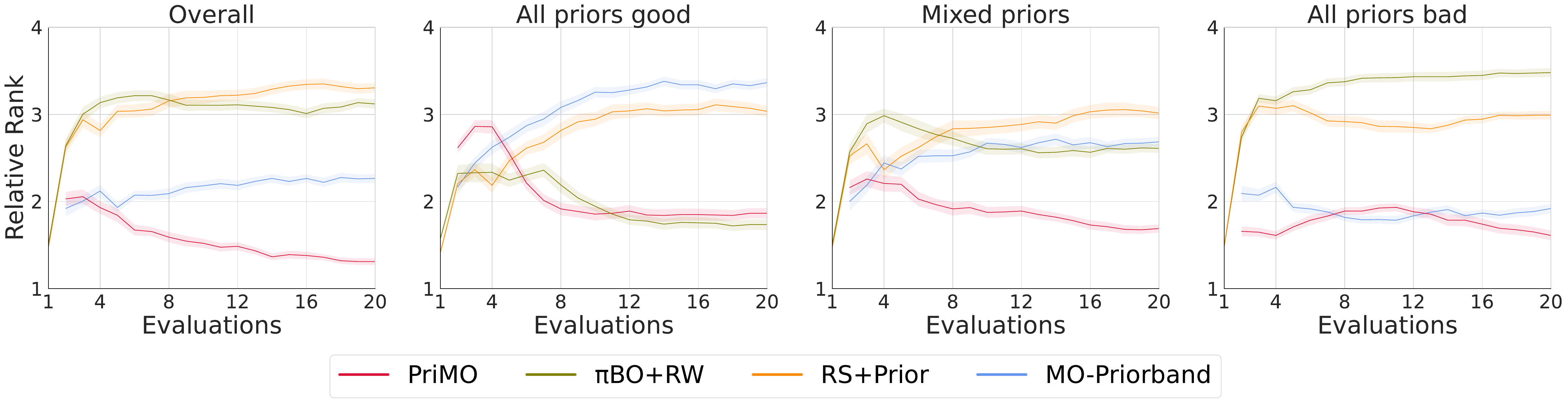}
    \caption{Mean relative ranks $\pm$ 1 standard error of baselines we constructed to use multi-objective priors and \algo{} across benchmarks and seeds under various prior conditions.}
    \label{fig:ovrank_subset2_all_4}
\end{figure*}

In this section, we present the relative rankings and detailed \hvlong{} plots across all our 8 benchmarks, comparing \algo{} against some prior-based baselines, adapted by us to the \mo{} case.
\Figref{fig:ovrank_subset2_all_4} clearly shows that, overall, \algo{} is anytime better compared to prior-based \mo{} adapted baselines and outperforms \mopb{} by a wide margin. Under good priors, \algo{} and \piborw{} are usually 2 of the best optimizers on average. 
This is expected in terms of the \allg{} priors setting since \algo{}'s \bo{} phase is based on \pibo{}'s design.

\begin{figure}[tb]
    \centering
    \includegraphics[width=1\linewidth]{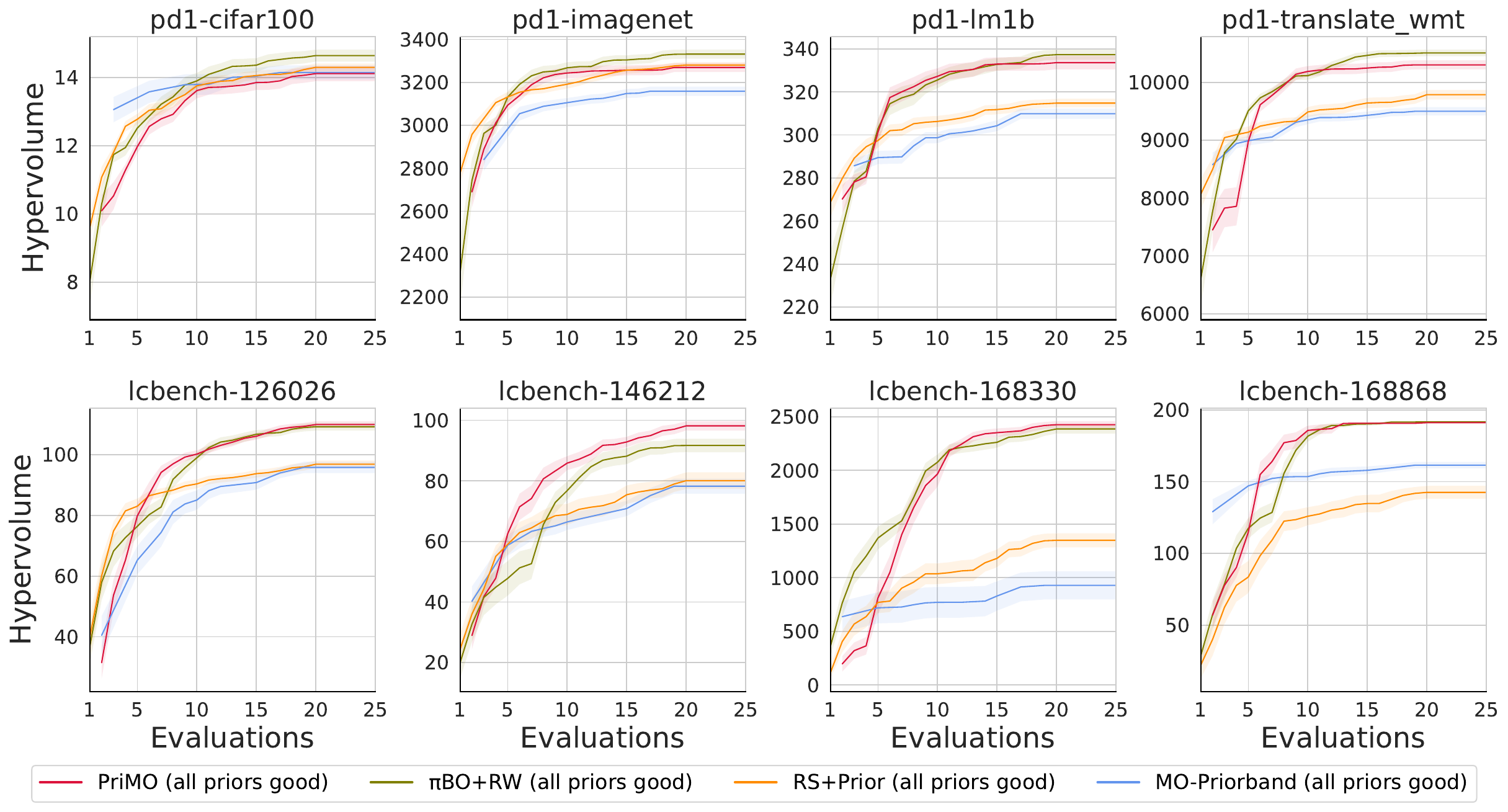}
    \caption{Comparing the average dominated \hvi{} over \nfevals{} evaluations and across 25 seeds, between \algo{} and prior-based baselines \mopb{}, \piborw{} and \RSprior{}, under all good prior conditions.}
    \label{fig:hv_subset2_all_good}
\end{figure}

The plots in \Figref{fig:hv_subset2_all_good} demonstrate that, with the exception of the \cifar{} benchmark, \algo{} is one of the two best optimizers across all benchmarks. This highlights \algo{}'s ability to effectively utilize good priors despite the \eps{}-greedy non-prior based component of its \bo{}. \piborw{}is marginally better than \algo{} in some benchmarks, due to its much longer dependence on the priors. On the other hand, \mopb{} seems to be quite ineffective in the utilization of good priors and is the worst performing \hpo{} algorithm across most benchmarks.


\subsection{Robustness of PriMO under bad prior conditions}\label{app_sec:res_robust}

\begin{figure}[tb]
    \centering
    \includegraphics[width=1\linewidth]{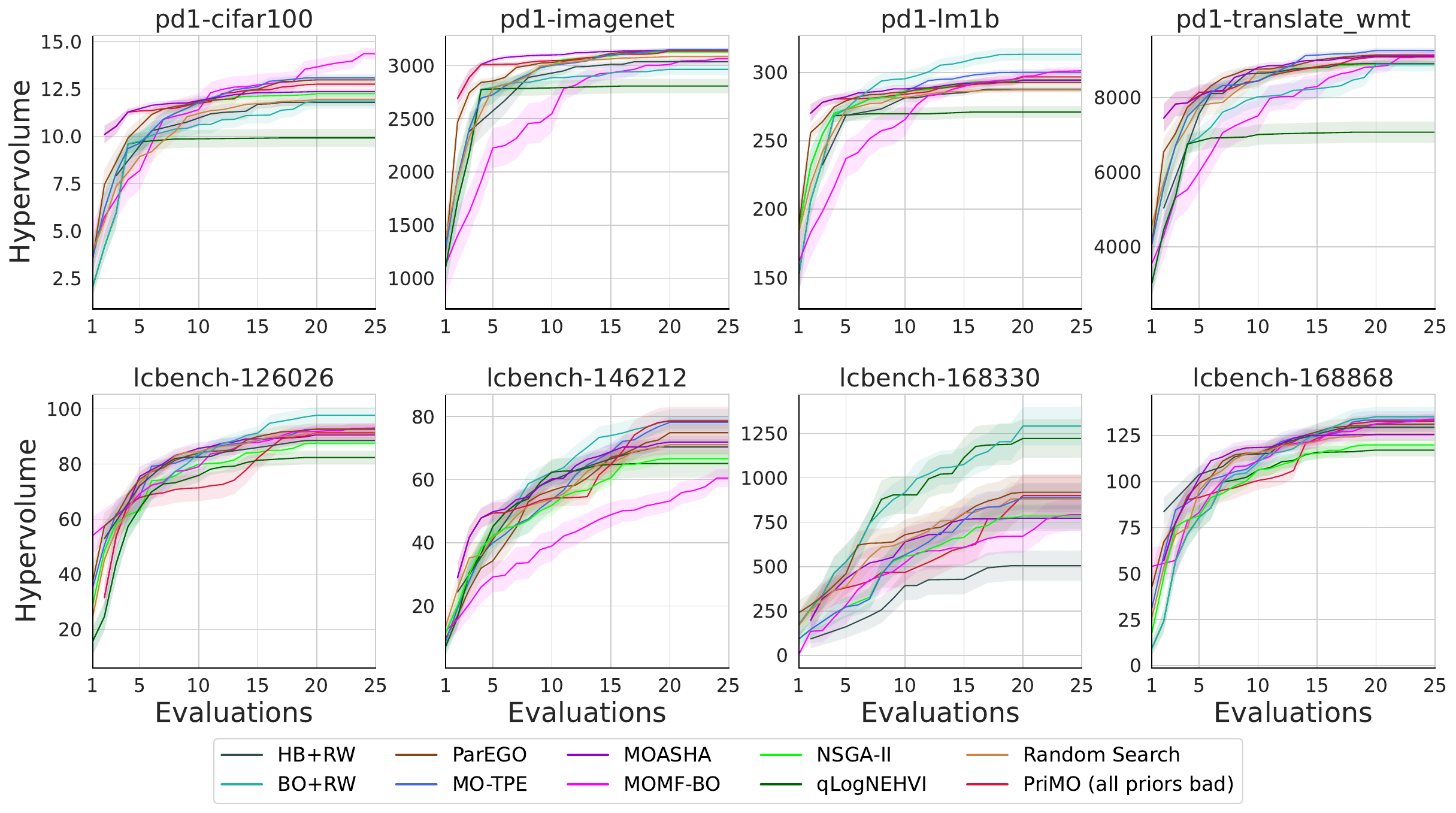}
    \caption{Mean dominated \hv{} across 8 Deep Learning benchmarks, showcasing remarkable recovery of \algo{} from bad priors, compared against some prominent non-prior baselines.}
    \label{fig:hv_subset1_all_bad}
\end{figure}

\Figref{fig:hv_subset1_all_bad} shows \algo{}'s remarkable ability to recover from bad priors with respect to the dominated \hvlong{}, across all benchmarks. At the end of the optimization budget, \algo{} achieves a competitive final performance on most benchmarks.

\begin{figure}[tb]
    \centering
    \includegraphics[width=1\linewidth]{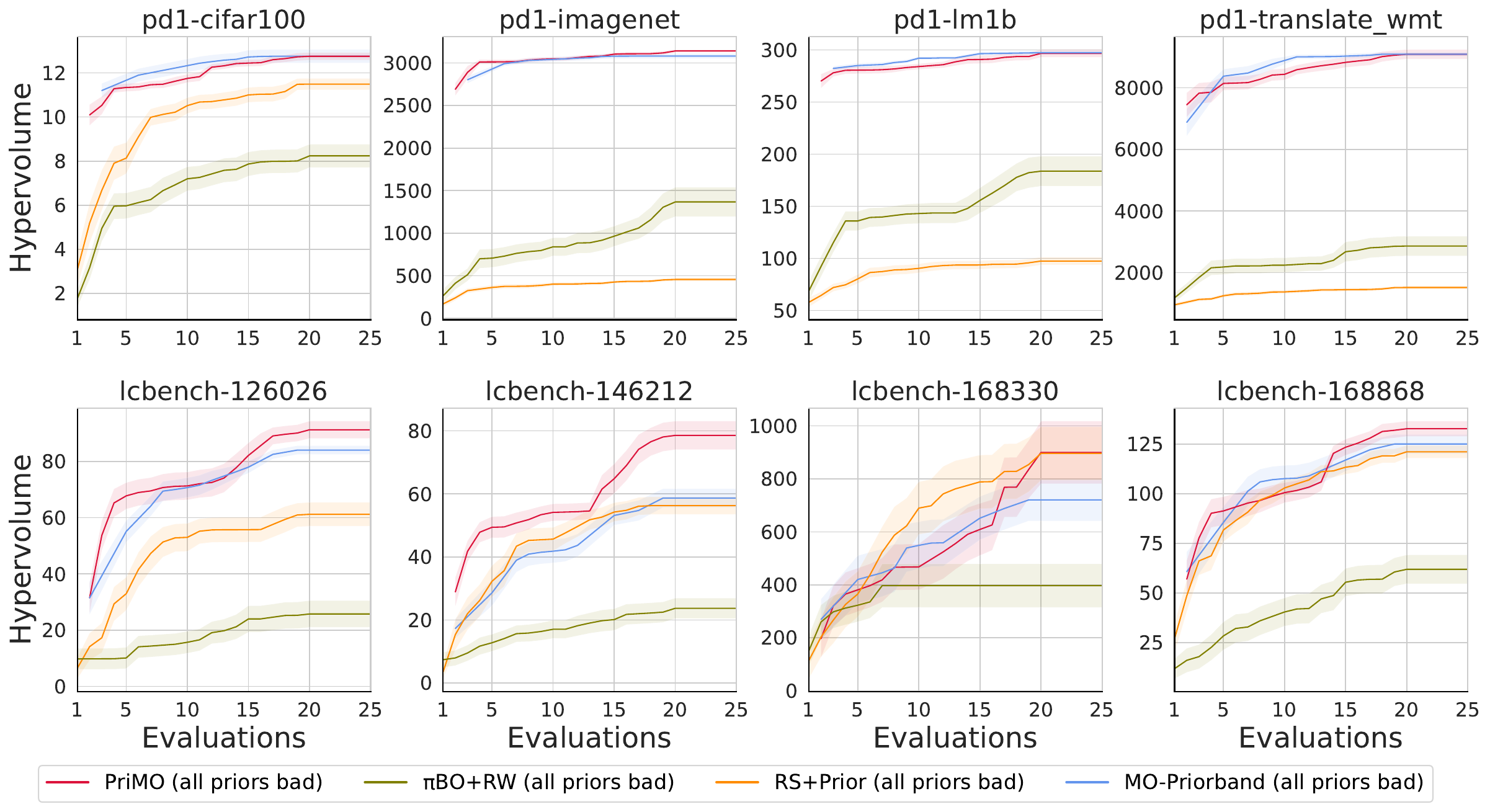}
    \caption{Dominated \hv{} plot comparing \algo{} against prior-based baselines that were adapted to \mo{}, under bad priors.}
    \label{fig:hv_subset2_all_bad}
\end{figure}

Compared to \mo{}-adapted prior-based baselines in \Figref{fig:ovrank_subset2_all_4} (\allb{}), we observe that \algo{} not only has good anytime performance, but also significantly outperforms all baselines by the end of the optimization run across most benchmarks. Thus, \piborw{}, despite being able to leverage good priors well, is quite prone to misleading priors and, overall, significantly (\appref{app:sig_analysis}) less robust compared to \algo{}.
This performance is further detailed in the \hvlong{} plots in \Figref{fig:hv_subset2_all_bad}. \algo{} is clearly shown to be the best performing optimizer across all benchmarks, with \mopb{} - a close second. \piborw{} is unable to recover from bad priors and is the algorithm with the worst final performance on most benchmarks.

We attribute \algo{}’s strong recovery under misleading priors to our design of the \textit{decaying \mo{}-prior-weighted acquisition}, influenced by two key parameters - $\beta$ and $\epsilon$. 
Unlike \piborw{}, which relies on the prior for much longer due to its slow decay schedule, \algo{} is explicitly designed to reduce prior influence more aggressively. This design choice allows \algo{} to recover more quickly when the priors are misleading, whereas \piborw{}’s prolonged dependence on bad priors significantly hinders its performance, resulting in noticeably worse final performance compared to both \algo{} and \mopb{}. 
The aggressive $\beta$ setting ensures the prior’s influence diminishes rapidly — an important property for practical \dl{} scenarios where \hpo{} is not expected to be run for long. Additionally, the parameter $\epsilon$ in the acquisition function controls how much the prior contributes while it is still active, thus encouraging exploration of the search space. Together, these two effects ensure that \algo{} does not become overly dependent on the prior and, under inaccurate priors, can still effectively explore and discover better \hp{} configurations than its counterparts.


\subsection{A note on compute efficiency}
\begin{figure}[tb]
    \centering
    \includegraphics[width=1\linewidth]{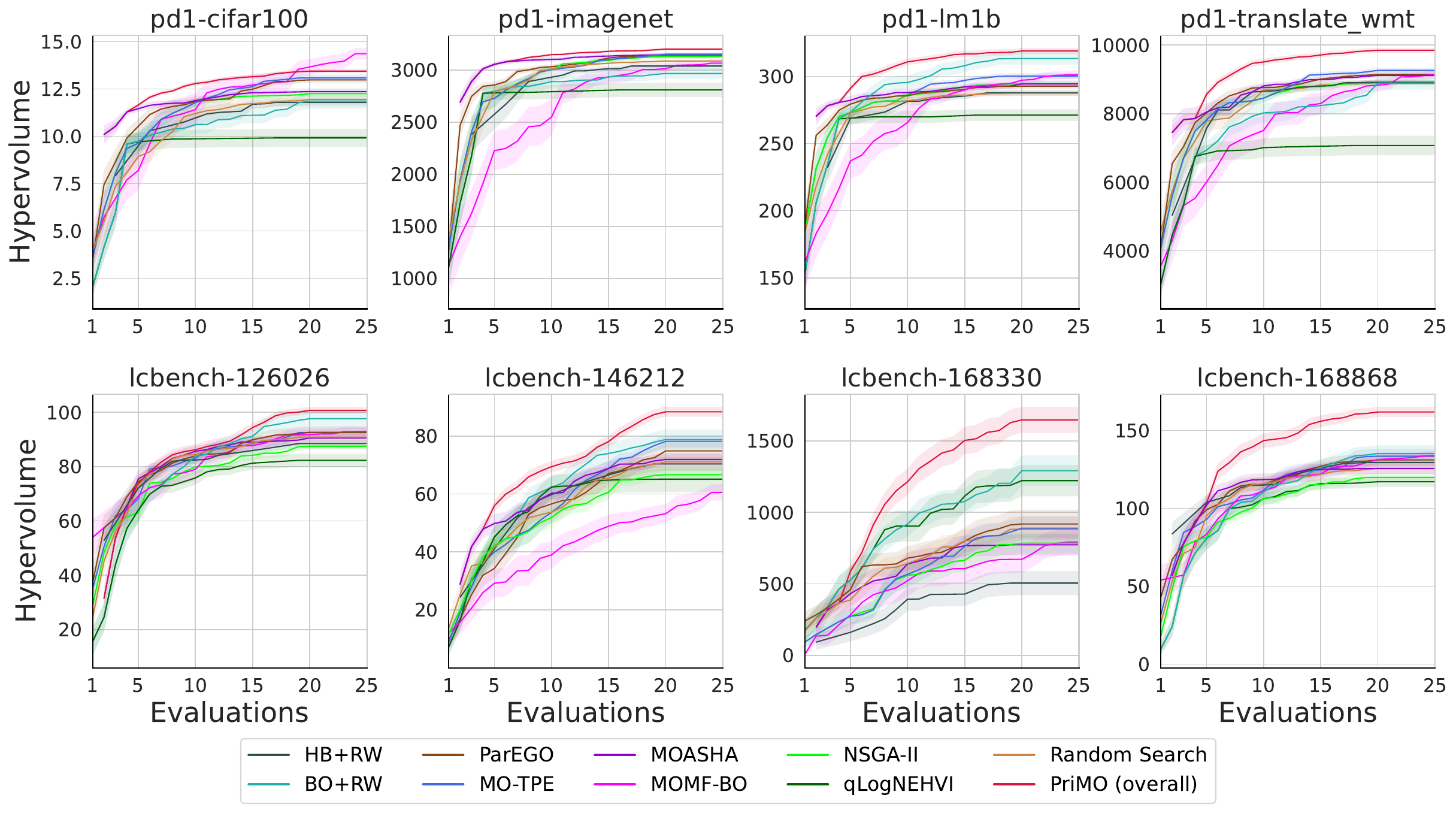}
    \caption{Comparing \algo{} against some prominent non-prior \mo{} baselines under the \ova{} prior setting, with respect to the mean dominated \hv{} across all 8 benchmarks.}
    \label{fig:hv_subset1_overall}
\end{figure}

\algo{} stands out as being \textit{extremely compute-efficient}, on average, achieving significant performance gains with minimal \hpo{} evaluations, \ie{}, with a low compute budget. We study this under \ova{} prior conditions with respect to the dominated \hv{} in \Figref{fig:hv_subset1_overall}. Given that we set \algo{}'s \init{} size to 5, an asynchronous \mf{} optimizer like \moasha{} (in a continual setup) effectively requires only about 3.5 equivalent function evaluations, which on average results in 3 configurations sampled at $\maxfid{}$. Therefore, compared to other \bo{} algorithms whose \init{} size we set to the number of dimensions, \algo{} effectively uses fewer max-fidelity configurations to fit the \gp{} in the \bo{} phase. Despite fewer samples, \algo{} already achieves much better performance in the beginning compared to all \bo{}-based baselines on most benchmarks, due to the use of its \init{} strategy.

In summary, these findings support our claim that \algo{} is a robust and general purpose \mohpolong{} algorithm designed for real-world \dl{} workloads, fulfilling all the desiderata outlined in \tabref{table:desiderata}. 

\section{Significance analysis}\label{app:sig_analysis}

In this appendix, we perform statistical significance tests using Linear Mixed Effect Models (LMEMs) to verify the results obtained in our experiments. Our choice for using LMEMs is supported by \citet{riezler_validity_2022} who proposed LMEM-based significance testing for Natural Language Processing tasks. Further, \citet{geburek2024lmemsposthocanalysishpo} argued for the usage of LMEM-based significance analysis for \hpo{} benchmarking.

\subsection{Data preparation and sanity checks}
To prepare the data for the significance analysis, we computed and used normalized \hvlong{} regret scores, as the scale of \hv{} can vary considerably across benchmarks. After aggregating the normalized \hv{} regret values at each function evaluation, we conducted sanity checks to ensure statistical validity. We then performed a post-hoc analysis and used Critical Difference (CD) diagrams to compare the early and final performance of \algo{} against all other algorithms.

\paragraph{Seed independency check}
We fitted two LMEMs:
\begin{equation}
\texttt{normalized\_hv\_regret $\sim$ algorithm} \quad,
\end{equation}
and,
\begin{equation}
\texttt{normalized\_hv\_regret $\sim$ (0 + algorithm | seed)} \quad,
\end{equation}
on the data and performed a Generalized Likelihood Ratio Test (GLRT) to verify that the seed is not a significant effect.

\paragraph{Benchmark informativeness}

Using GLRT, we compared the likelihoods of the LMEMs: 
\begin{equation}
\texttt{normalized\_hv\_regret $\sim$ 1} \quad,
\end{equation}
and,
\begin{equation}\label{eq:anova}
\texttt{normalized\_hv\_regret $\sim$ algorithm} \quad,
\end{equation}
which confirmed that our benchmarks are informative, as the second model (Equation \ref{eq:anova}) was shown to be significantly better. This further indicates that there are indeed significant differences between the performance of algorithms across all benchmarks, justifying the use of CD diagrams for comparison.

\subsection{Critical difference diagrams}

We perform pairwise Tukey HSD \citep{Tukey1949ComparingIM} tests using LMEMs to obtain individual p-values for each comparison. Using this, we plot the CD diagrams.

\begin{figure}
    \centering
    \includegraphics[width=1\linewidth]{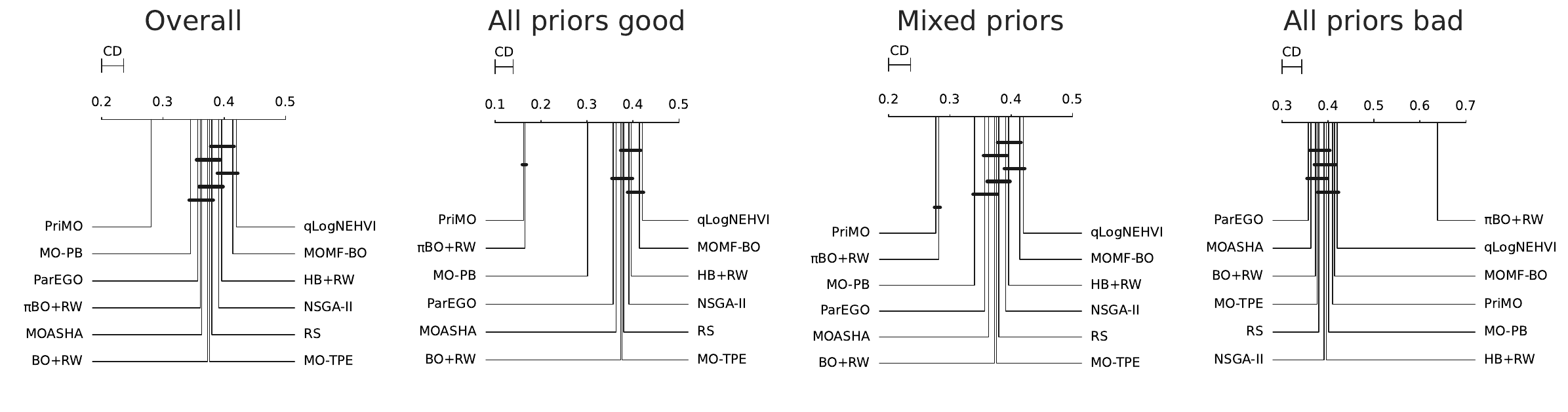}
    \caption[Critical Difference diagrams at 10 function evaluations under various prior conditions]{Critical Difference diagrams at \textbf{10 evaluations} comparing \textbf{early performance} of \algo{} against the baselines -- \borw{}, \piborw{}, \mopb{} (MO-PB), \moasha{}, \hbrw{}, \motpe{}, \lnehvi{}, \RSlong{} (\RS{}), \parego{} and \nsgaii{}, under various prior conditions.}
    \label{fig:cd_plots_iter10_all}
\end{figure}

Here, we consider the statistical differences in the early and final performance between \algo{} and other algorithms. \Figref{fig:cd_plots_iter10_all} shows CD plots for 10 function evaluations, \ie{}, halfway through our entire allocated budget. \Figref{fig:cd_plots_iter10_all} (\allg{}) shows that \algo{} is able to efficiently leverage good priors very early during the optimization, and significantly better than all baselines except \piborw{}. Under all bad priors, \algo{}'s performance is not significantly worse than that of most the of best performing non-prior baselines, except \borw{}, but is significantly better than \piborw{}. However, averaging all prior conditions in \figref{fig:cd_plots_iter10_all} (\ova{}), we observe a significant difference between \algo{} and all other optimizers. \algo{} is shown to be the best ranked algorithm with significantly strong early performance.

\begin{figure}
    \centering
    \includegraphics[width=1\linewidth]{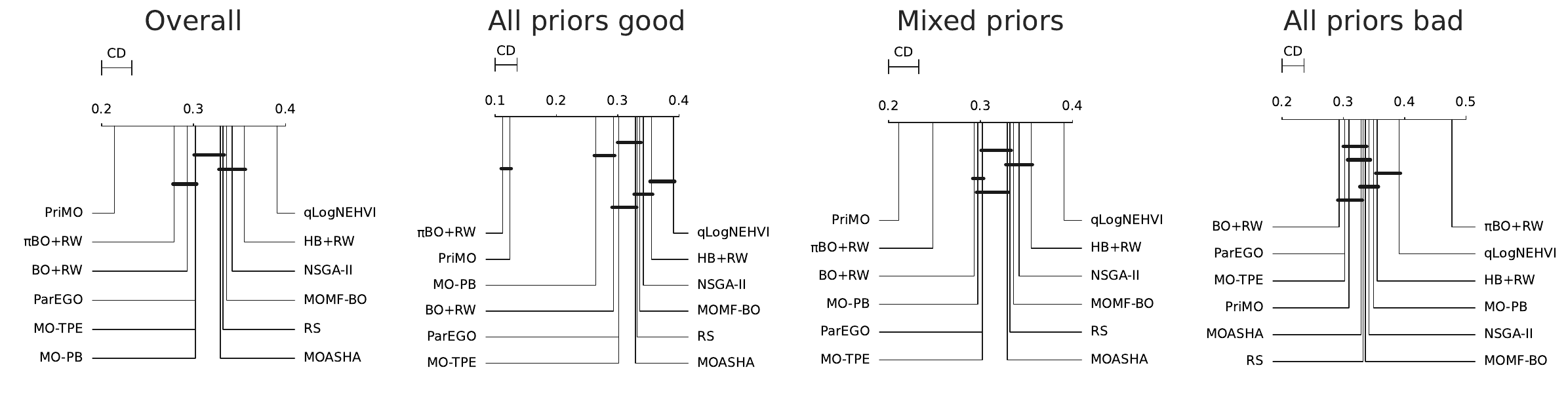}
    \caption[Critical Difference diagrams at 20 function evaluations under various prior conditions]{CD diagrams at \textbf{20 evaluations} comparing \textbf{final performance} of \algo{} against other non-prior and prior-guided (adapted to \mo{}) baselines, under various prior conditions.}
    \label{fig:cd_plots_iter20_all}
\end{figure}

In \Figref{fig:cd_plots_iter20_all}, we show the CD diagrams for 20 function evaluations, \ie{}, at the end of our optimization budget. As observed in our relative ranking plots before, there is negligible critical difference between \algo{} and \piborw{} under all good priors, while both are significantly better than other algorithms. \Figref{fig:cd_plots_iter20_all} (\allb{}) verifies the final performance of \algo{} under bad priors, highlighting a strong recovery, where, with the notable exception of \borw{}, \algo{} is shown to be significantly better than all baselines. While not significantly better than \borw{} under \mxd{}, \algo{} is nevertheless the best ranked optimizer. \ova{}, \algo{} is clearly shown to be the algorithm with the highest rank, indicating the strongest final performance. 

Thus, Figures~\ref{fig:cd_plots_iter10_all} and \ref{fig:cd_plots_iter20_all} confirm our relative ranking plots and statistically verify \algo{}'s \sota{} performance, proving that overall, \algo{} is significantly better than all other algorithms used in our study.






\section{Code repository}\label{app_sec:repo}


The $\texttt{Python}$ code for generating the priors and running the experiments presented in this paper is publicly available in \href{https://anonymous.4open.science/r/momfpriors_anon-7AE5}{$\texttt{this}$} repository. This repository also contains the code used to generate all plots, along with a comprehensive $\texttt{README.md}$ file that provides reproducibility guidelines, explains the output data structure, and outlines the steps required to run all baselines on the benchmarks used in this work. The priors over the objectives for the various benchmarks, and the raw results from the all optimization runs of \algo{} used in this paper are also included.








\section{Resources used}\label{app:resources}

We ran all the algorithms in this paper on inexpensive surrogate and synthetic benchmarks. To perform all our experiments, we only used a CPU compute cluster and 30 cores of Intel(R) Xeon(R) Gold 6242 CPU @ 2.80GHz. For runs up to \nfevals{} function evaluations, each seed of an \hpo{} algorithm on a single benchmark took $\sim$0.025 CPU hours, or $\sim$0.75 core hours on average. While \mf{} optimizers such as \moasha{} and \hbrw{} completed in just a few seconds ($\sim$0.15 core hours), model-based baselines such as \borw{} and \piborw{} required significantly longer on average - typically over 5 minutes ($\sim$2.5 core hours).

For the experiments in \Secref{sec:exp}, we ran 19 optimizers in total - 12 non-prior and 9 prior-based, including \algo{} and all its design ablations, and the optimizers in the single-objective setting. Each prior-based \molong{} optimizer was evaluated under 4 different prior combinations, whereas 2 priors were used for the prior-based \solong{} optimizers. Each run lasted \nfevals{} evaluations and we evaluated each optimizer on 8 benchmarks across 25 seeds. In total, this amounted to $\sim$208 CPU hours, or $\sim$6240 core hours to generate the results presented in \Secref{sec:exp}.




\section{Licenses}\label{app:licenses}


\begin{itemize}
    \item NePS package: \textbf{Apache License, Version 2.0}
    \item \hpogluett{}: \textbf{BSD 3-Clause License}
    \item $\texttt{mf-prior-bench}$: \textbf{Apache License, Version 2.0}
    \item \yahpogym{}: \textbf{Apache License, Version 2.0}
    \item \hyperbo{} \pdone{}: \textbf{Apache License, Version 2.0}
    \item Nevergrad: \textbf{MIT License}
    \item SMAC: \textbf{BSD 3-Clause License}
    \item Syne Tune (code for \epsnet{}): \textbf{Apache License, Version 2.0}
    \item $\texttt{lmem-significance}$: \textbf{MIT License}
\end{itemize}

\end{document}